\documentclass[12pt,ijoc,nonblindrev]{article}

\PassOptionsToPackage{numbers, compress}{natbib}
\usepackage[preprint]{neurips}

\usepackage{amsmath,amssymb,amsfonts}

\usepackage{natbib}
 \bibpunct[, ]{(}{)}{,}{a}{}{,}%
 \def\bibfont{\small}%

\usepackage{rotating}
\usepackage{fancyvrb}
\usepackage{mathtools}
\usepackage{bbm}

\newenvironment{proofsketch}{{\bf \emph{Proof Sketch.} }}{\hfill $\Box$} 

\newcommand{\parameter}{\omega}

\newcommand{\med}{\parameter}

\usepackage{wrapfig}
\usepackage{subfigure}
\usepackage{algorithm2e} 
\usepackage{url}

\usepackage[dvipsnames]{xcolor}

\newtheorem{assumption}{Assumption}
\newtheorem{proposition}{Proposition}
\newtheorem{definition}{Definition}
\newtheorem{lemma}{Lemma}
\newtheorem{remark}{Remark}
\newtheorem{algorithms}{Algorithm}
\newtheorem{theorem}{Theorem}

\ifodd 1
\newcommand{\com}[1]{\textbf{\color{red}(Parinaz's comment: #1)}} 
\newcommand{\comy}[1]{\textbf{\color{cyan}(Yifan's comment: #1)}} 
\newcommand{\res}[1]{\textbf{\color{magenta}(RESPONSE: #1)}} 
\newcommand{\yl}[1]{\textbf{\color{red}(Yang's comment: #1)}} 

\else

\newcommand{\com}[1]{}
\newcommand{\comy}[1]{}
\newcommand{\yl}[1]{}
\newcommand{\res}[1]{}
\fi
\title{Adaptive bounded exploration and intermediate actions for data debiasing}

\author{%
        Yifan Yang \\ 
        Ohio State University \\
        \texttt{yang.5483@osu.edu} \\
        \And Yang Liu \\ 
        UC, Santa Cruz \\
        \texttt{yangliu@ucsc.edu} \\
        \And Parinaz Naghizadeh \\
        UC, San Diego \\
        \texttt{parinaz@ucsd.edu} \\  
}

\begin{document}

\maketitle

\begin{abstract}
The performance of algorithmic decision rules is largely dependent on the quality of training datasets available to them. Biases in these datasets can raise economic and ethical concerns due to the resulting algorithms' disparate treatment of different groups. In this paper, we propose algorithms for sequentially debiasing the training dataset through adaptive and bounded exploration in a classification problem with costly and censored feedback. Our proposed algorithms balance between the ultimate goal of mitigating the impacts of data biases -- which will in turn lead to more accurate and fairer decisions, and the exploration risks incurred to achieve this goal. Specifically, we propose adaptive \emph{bounds} to limit the region of exploration, and leverage \emph{intermediate actions} which provide noisy label information at a lower cost. We analytically show that such exploration can help debias data in certain distributions, investigate how {algorithmic fairness interventions} can work in conjunction with our proposed algorithms, and validate the performance of these algorithms through numerical experiments on synthetic and real-world data. 
\end{abstract}

\section{Introduction}\label{sec:intro}

Data-driven algorithms are used to guide or make decisions in various application domains with considerable impact on human subjects, including loan approvals, legal recidivism assessment, and allocation of medical resources. Despite their high prediction accuracy and scalability, there have also been concerns regarding these algorithms' potential negative (social) impacts, as it has been observed that these algorithms may unintentionally amplify existing social biases \citep{dressel2018accuracy, lambrecht2019algorithmic, obermeyer2019dissecting}. The observed unfairness in these decision systems may be due to data biases and/or algorithmic issues~\citep{mehrabi2021survey}. In this paper, we focus on the former issue of statistical biases in the training data. We propose \emph{adaptive data debiasing} algorithms to {mitigate existing training data biases and guide future data collection}, ultimately leading to both more accurate and fairer decisions across diverse demographic groups, thereby promoting social good.

In particular, the datasets used to train machine learning algorithms might not accurately reflect the characteristics of the populations impacted by the algorithm's decisions. This mismatch can stem from historical biases in decision-making, erroneous feature selection and measurement, data labeling errors, or changes in population characteristics after initial data collection. These data biases can result in not only sub-optimal (low accuracy) decisions, but also the disparate treatment of underrepresented groups~\citep{kallus2018residual, wang2021fair, zhu2021rich}, even if algorithmic fairness interventions are implemented~\citep{liao2023social}. In other words, the literature has repeatedly identified statistical biases in training datasets as a source of algorithmic unfairness, as well as a hindrance to the application of fairness interventions to prevent algorithms from suggesting discriminatory decisions for different demographic groups. For instance, statistical biases in the training data may lead a financial institution to erroneously issue loans to unqualified applicants, or to (inadvertently) violate the Equal Credit Opportunity Act; or, an employer using an automated resume screening software may not be able to select the most qualified applicants for interviews, while even discriminating in its selection process based on their race or gender. As such, decision makers have an incentive to  mitigate data biases to ensure high accuracy (and hence, profit), as well as to maintain business reputation and meet any applicable legal requirements. Motivated by this, our work proposes a \emph{data debiasing} algorithm that can guide the collection of new data over time so that training data more closely matches the true population statistics, ultimately leading to more accurate and fairer decisions. 

Specifically, we study a classification problem in which the decision maker has access to an initially (statistically) biased training dataset, and additionally, faces \emph{censored and costly feedback} when collecting future data. \emph{Censored feedback} means that the true label (qualification state) of an individual will be revealed to the decision maker only if that individual is labeld positively. For instance, banks can ascertain whether a loan recipient defaults or repays only after extending the loan; or, an employer can only evaluate the performance of applicants that are ultimately hired. In these and other application areas (e.g., school admissions, recidivism decisions), once a decision rule has been selected, future data is restricted to those meeting the current approval criteria. As such, the algorithm will only have restricted access to the data domain, and its training data will grow in a biased way going forward. One way to address this issue is for the decision makers to ``explore'': go against the algorithm's recommendation with {the goal of collecting otherwise unobserved data}, albeit at a cost (e.g., extending loans to or hiring potentially unqualified individuals); this is the \emph{costly} nature of feedback.

The idea of using (pure) exploration to overcome censored feedback has been considered in some recent works \citep{bechavod2019equal,kilbertus2020fair,wei2021decision}. Our proposed \texttt{Active Debiasing} algorithm also uses exploration to mitigate data biases, but unlike existing works, also limits its costs through \emph{bounded exploration}, strategically admitting some agents that would otherwise be rejected, while adaptively restricting the extent and frequency of this exploration. Specifically, our algorithm includes two parameters to limit exploration costs: one modulates the \emph{frequency} of exploration (an exploration probability $\epsilon_t$, common in the (reinforcement/bandit) learning literature, which can be adjusted using current bias estimates), and another limits the \emph{depth} of exploration (by setting a lower bound ${LB}_t$, a new idea in our algorithm, on how far the decision maker is willing to deviate from the current optimal policy when exploring). This new exploration lower bound poses challenges in proving that our algorithm can recover unbiased estimates of the unknown population statistic. Specifically, the proof involves the analysis of statistical estimates $\hat{\omega}_t$ based on data collected from \emph{truncated} distributions due to the use of an exploration lowerbound, from an exploration range $[{LB}_t , \infty)$ that is itself adaptive. To address this challenge, we first analyze the sequence of estimates in finite sample regimes, and then prove that the sequence of over- and under- estimation errors converge to zero-mean random variables with variance going to zero as the number of samples increases in {unimodal} distributions (Theorem~\ref{thm:debiasing}). Together, these establish that our proposed algorithm can mitigate the impacts of initial biases in the training data, and further prevent additional biases due to censored feedback.

Building on these insights, we next note that existing works on using exploration to overcome censored feedback have only considered \emph{binary} exploration options. For instance, in the lending scenario, this means that a financial institution can either approve or deny a loan application; however, in practice, while a large loan may be denied (if a bank perceives it to be high risk), a smaller loan may still be extended, offering a (noisy) evaluation of the applicant's qualification for the larger loan. Similarly, current methods assume that an employer only has one (full-time) hiring option; in practice, the employer may consider short-term contracts or internships to obtain a (potentially noisy) assessment of an individual's performance on the job. Inspired by these, we introduce an additional \emph{intermediate} exploration action in the data debiasing problem, through which the algorithm can obtain a noisy observation of an agent's true label at a lower cost. While offering an intermediate action to some agents during exploration (as opposed to offering a \emph{uniform} action to all explored agents) could lower costs, one might expect that the noisy label information will slow the rate of debiasing. Therefore, to formally investigate the trade-off between the debiasing speed and the incurred costs when employing intermediate actions, we consider the problem of making intermediate vs. uniform exploration decisions in a two-stage Markov Decision Process (MDP) framework. We use this model to show that the debiasing speed is faster with uniform exploration when compared to intermediate action, albeit at the expense of higher incurred costs under certain conditions. The main challenge is to evaluate the impact of the explored data quality (affected by the noisy nature of intermediate actions) on mitigating the data bias when employing the \texttt{active debiasing} algorithm. Specifically, to address this challenge, we first analytically solve for the decision threshold in terms of the feature-label distribution estimates, and then show how a better quality of data could help the debiasing procedure.  

To summarize, our main findings and contributions are:

1. {\emph{A bounded exploration algorithm, with analytical support.} We propose an adaptive data debiasing algorithm with bounded exploration to effectively mitigate statistical biases in the training data in classification problems with censored feedback, while limiting exploration costs. We analytically show that our proposed algorithm can mitigate biases in unimodal distribution estimates (Theorem~\ref{thm:debiasing}). We also provide an error bound (on the number of wrong decisions) for our algorithm (Theorem~\ref{thm:regret}), and its impacts when used in conjunction with existing algorithmic fairness interventions (Proposition~\ref{prop:fairness-debiasing}).} 

2. {\emph{Beyond binary exploration decisions.} We then propose the use of intermediate exploration actions to further reduce the costs of exploration. We model the exploration decision processes as a two-stage MDP, and analyze the impact of opting for intermediate actions on the trade-off between debiasing speed and cumulative costs (Theorems~\ref{thm:theta} and \ref{thm:mdp})}.

3. \emph{Numerical experiments.} We provide numerical support for our proposed algorithms through experiments on synthetic and real-world datasets (the Adult dataset~\citep{Dua:2019}, {the Retiring Adult dataset ~\citep{ding2021retiring}}, and a FICO credit score dataset~\citep{hardt2016equality}). We show that our algorithms can successfully mitigate data biases, leading to both more accurate and fairer algorithmic decisions.

The remainder of this paper is organized as follows. Section~\ref{sub:illustaritve} provides an illustrative example to introduce the problem background. Section~\ref{sec:related} provides a review of related work. Section~\ref{sec:model} introduces our model framework and some preliminaries. Our proposed \texttt{active debiasing} algorithm is detailed in Section~\ref{sec:single-group}, followed by a theoretical analysis of its properties in Section~\ref{sec:analytical}. Section~\ref{sec:mdp} goes beyond the binary decision and evaluates the intermediate exploration action.
Numerical experiments are presented in Section~\ref{sec:simulations}.
We conclude with a discussion of limitations and potential future work in Section~\ref{sec:conclusion}.

\subsection{Illustrative example: loan application decisions based on credit scores}\label{sub:illustaritve}

As users' understanding of scoring systems and their access to resources evolve continuously over time, it has been found that the average U.S. FICO credit score is trending upward \citep{avg-CS, avg-CS-2}. At the same time, it has been found that Black and Hispanic applicants have, historically, had lower credit scores than their counterparts in White applicants, and therefore it has been argued that relying on these historically low credit scores can perpetuate the cycle of discrimination against minority groups \citep{black-white}. Consequently, if financial institutions continue to rely on historical data and static decision thresholds based on past credit scores to issue loans or credit cards to applicants, their decisions may become suboptimal due to either shifts in the FICO score distribution, or historical biases in them. 

To address these data bias issues and overcome the challenge of censored feedback in data collection — where true label information (i.e., whether an individual will default) is only available when a loan is granted (a positive decision) — financial institutions need to collect more diverse and higher-quality samples to assess different individuals' credit scores vs. eventual creditworthiness, and update their selection criteria (e.g., thresholds on credit scores for approving applicants for a loan or a credit card) accordingly. 

If decision-makers persist with the current (and potentially biased) threshold by only approving applicants above a certain credit score for loans, without exploration, this approach can lead to setting thresholds that are too high (as we find based on our model in Theorem~\ref{thm:exploit-only_and_pure-exploration}), leading to under-selection of minority groups. In the long run, this could damage the institution's reputation and incur significant losses, and perpetuate discrimination across different demographic groups. Thus, decision-makers need to incorporate exploration, going against their current algorithm's recommendations, to mitigate bias arising from the data change. Exploration, however, incurs costs, as applicants below the current decision threshold can not be freely assessed (e.g., issuing loans to potentially unqualified individuals leads to a loss for the financial institution).

To limit the costs in new data acquisition, we propose that these firms should, as a first step, adopt a bounded exploration technique (as outlined in Section~\ref{sec:single-group}) by introducing an exploration lower bound (as defined in Definition~\ref{def:UB-LB}). This means that the institution should \emph{sometimes} extend loans to a few of the applicants that are currently not qualified to receive a favorable decision, but to \emph{limit} this to applicants who still pass a minimum ($LB$) criteria to limit risks. This approach manages exploration costs by controlling both the depth of exploration ($LB$ in our model) and the frequency of exploration ($\epsilon$ in our model). Additionally, institutions may further improve their exploration strategies by refining their exploration decisions' categories. In particular, a financial institution may note that different exploration decision types have substantially different costs (e.g., issuing loans to unqualified individuals is far more costly than rejecting qualified ones). Motivated by this, we explore ways to further minimize costs by considering different decision types within the exploration range, extending beyond binary decisions of granting or rejecting a loan (as discussed in Section~\ref{sec:mdp}). Specifically, decision-makers may opt for ``intermediate'' exploratory actions (such as issuing micro-loans or lower limit cards, before approving a major loan or increasing the credit limit), which are less costly but provide \emph{noisy} information about unobserved qualification states (e.g., an unqualified individual might repay a micro-loan on time, but still default on a larger loan). We show that despite the noisy nature of these labels, a financial institution may collect informative data in this way, to improve the quality of its decision in the long-run. We establish that this data debiasing strategy not only ultimately improves the firm's profit while limiting its costs of data collection, but more importantly, also leads to fairer and more equitable decisions for applicants from different demographic groups.

\section{Related work}\label{sec:related}

Our paper is most closely related to the works of \citep{ensign2018runaway,bechavod2019equal,kilbertus2020fair,blum2020recovering,jiang2020identifying}, which investigated the impacts of data biases on (fair) algorithmic decision-making under censored feedback. \citet{ensign2018runaway} were among the first to identify feedback loops between predictive algorithms and biases in the data used for training. However, their work does not address the impact of the distribution shift. In contrast, while both their work and ours study the effects of censored feedback on predictive algorithms, we focus on developing a debiasing algorithm that emphasizes the debiasing process itself and examines its interaction with fairness-constrained learning. \citet{bechavod2019equal} start with a pure exploration phase and subsequently refine their exploration to ensure that fairness constraints are upheld, while \citet{kilbertus2020fair} employ (pure) exploration strategies to address censored feedback. Although our work also adopts exploration strategies, the form and purpose of exploration differ significantly. We start with a {biased} dataset, and conduct \emph{bounded} exploration with the goal of data debiasing while accounting for the costs of exploration; fairness constraints may or may not be enforced separately and are orthogonal to our debiasing process. As shown in Section~\ref{sec:simulations}, such pure exploration processes incur higher exploration costs than our proposed bounded exploration algorithm.

A number of other works, including  \citep{deshpande2018accurate,nie2018adaptively,neel2018mitigating,wei2021decision,chien2023algorithmic,harris2024strategic} have, similar to our work, explored the question of biases induced by a decision rule on data collection, particularly when feedback is censored. \citet{deshpande2018accurate} study the inference in a linear model with adaptively collected data. While both their work and ours investigate the impact of adaptive sampling bias, their focus is on debiasing an estimator, whereas our work concentrates on modifying the decision rule used for data collection. Similarly, \citet{nie2018adaptively} address the problem of estimating statistical parameters from adaptively collected data, proposing a random exploration technique based on data splitting and modified maximum likelihood estimators. In contrast, our \emph{bounded} exploration strategy explicitly considers the risks of exploration decisions and limits the depth of exploration to account for its costs. Unlike the ex-post debiasing methods for adaptively collected data proposed by \citep{deshpande2018accurate,nie2018adaptively}, \citet{neel2018mitigating} propose an adaptive data-gathering procedure, demonstrating that no debiasing is necessary if data is collected using a differentially private method. Similarly, we propose a debiasing algorithm that adaptively adjusts the data collection procedure. However, unlike \citep{neel2018mitigating}, our approach explicitly addresses the impact of distribution shift and the costs of exploration in the data collection process. \citet{wei2021decision} investigates data collection in the presence of censored feedback, considering the costs of exploration. They frame the problem as a partially observable Markov decision process, demonstrating that the optimal data collection policy is a threshold policy that becomes more stringent (in our terminology, reduces exploration) as learning progresses. While both their work and ours utilize adaptive and cost-sensitive exploration, we differ in our problem setup and our analysis of fairness constraints. More importantly, in contrast to both \citep{neel2018mitigating,wei2021decision}, our starting point is a \emph{biased} dataset (which may be biased for reasons other than adaptive sampling in its collection); we then consider how, while attempting to debias this dataset by collecting new data, any {additional} adaptive sampling bias during data collection should be prevented. \citet{chien2023algorithmic} examine the harm caused by selective labeling in dynamic learning systems, proposing a \emph{random} exploration method to collect samples that would otherwise be rejected. \citet{harris2024strategic} address censored feedback in scenarios where agents can strategically modify input features to obtain favorable predictions, leading to changes in the decision boundary. Their algorithm begins with pure exploration and later adjusts the decision boundary to collect clean, contextually relevant samples for updates while accounting for agent strategizing. In contrast to these approaches, although we also use exploration and adjust decision boundaries, the specific use of \emph{bounded} exploration, and accounting for fairness interventions are different from these works.

Our work is also related to the problem of data-driven inventory management with censored demand \citep{burnetas2000adaptive,godfrey2001adaptive,huh2009adaptive,huh2011adaptive,mersereau2015demand,agrawal2019learning,chen2020data,chen2021nonparametric,ding2024feature}. Some of these works also use a notion of exploration to address censored feedback \citep{burnetas2000adaptive,huh2011adaptive,mersereau2015demand,agrawal2019learning,chen2020data,chen2021nonparametric}. While the context of our work is different (inventory management vs. classification), our idea of bounded exploration and intermediate actions might be applicable in the inventory management context as well. Specifically, the decision maker typically observes only sales data and needs to (purely) explore larger inventory quantities (or, wider price range) to accurately estimate the demand distribution, which incurs high holding costs. Our \emph{bounded} exploration technique could be applied here and help to control the holding costs and other associated expenses. Additionally, the decision maker could offer ``Pre-Order'' services as an intermediate action to gather more demand data. This limits exploration costs, though the demand information might be noisy as customers could cancel their pre-orders afterward.

More broadly, our work is related to the literature on Bandit learning and its study of exploration and exploitation trade-offs, where adaptively adjusted exploration decisions play a key role in allowing the decision maker to attain new information, while at the same time using the collected information to maximize some notion of long-term reward. In particular, bandit exploration deviates from choosing the current best arm in several ways: randomly as in $\epsilon$-greedy, by some form of highest uncertainty as in Upper Confidence Bound (UCB) algorithm, by importance sampling approaches as in EXP3 algorithm, etc. A key difference of our work with these existing approaches is our choice of \emph{bounded} exploration, where the bounds are motivated by settings in which the cost of wrong decisions increase as samples further away from the current decision threshold are admitted. In that sense, our proposed approach can be viewed as a bounded version of $\epsilon$-greedy; we refer to the non-bounded version of $\epsilon$-greedy in our setting as \texttt{pure exploration}. 

A preliminary version of this work appeared in \citep{yang2022adaptive}, where we first introduced the idea of \emph{bounded} exploration for data debiasing. This paper extends \citep{yang2022adaptive} primarily by introducing the idea of intermediate actions and its corresponding analysis, as well as with extended numerical support, and by explicitly establishing the fairness benefits of bounded exploration. 

\section{Model and Preliminaries}\label{sec:model}

We consider a decision maker or firm, designing a data-driven algorithm to make decisions on a population of agents. Agent arrive over times $t=1, 2, \ldots$; the firm makes decisions on agents arriving at time $t$ based on it current algorithm, and adjusts the algorithm for times $t+1$ based on the observed outcomes.

Specifically, each agent has an observable \emph{feature} or \emph{score} $x\in \mathcal{X}\subseteq \mathbb{R}$ representing their characteristics (e.g., credit scores, exam scores). (We assume $\mathcal{X}\subseteq \mathbb{R}$ in our analysis, and generalize to $\mathcal{X}\subseteq \mathbb{R}^n$ in the experiments in  Section~\ref{sec:simulations}. We also discuss the implications of this assumption in Section~\ref{sec:conclusion}.) Agents are either qualified or unqualified to receive a favorable decision, denoted by their true label or qualification state $y\in \{0,1\}$. Additionally, agents belong to a group based on protected attributes (e.g., race, gender), labeled as $g\in \{a,b\}$. We consider threshold-based, group-specific, binary classifiers $h_{\theta_{g,t}}(x) =  \mathbbm{1}(x\geq \theta_{g,t})$ as the firm's adopted algorithm, where $\theta_{g,t}$ is the decision threshold. An agent from group $g$ with feature $x$ arriving at time $t$ is admitted iff $x \geq \theta_{g,t}$.

\textbf{Quantifying bias.} 
Let $f^y_{g}(x)=\mathbb{P}(X=x|Y=y, G=g)$ denote the true underlying pdf for the feature distribution of agents from group $g$ with label $y$. The algorithm builds an estimate $\hat{f}^y_{g,t}(x)$ of these unknown distributions at time $t$, based on the data collected so far (or an initial training dataset). In general, there can be a mismatch between the estimates $\hat{f}^y_{g,t}(x)$ and the true ${f}^y_{g}(x)$; this is what we refer to as {statistical data} bias. We make the following assumption to model this mismatch.

\begin{assumption}\label{as:distributions}
The firm updates its estimates $\hat{f}^y_{g,t}(x)$ by updating a single parameter $\hat{\parameter}^y_{g,t}$. 
\end{assumption}

This type of assumption is common in the multi-armed bandit learning literature \citep{schumann2022group,slivkins2019introduction,patil2021achieving,lattimore2020bandit,raab2021unintended}; there, the algorithm aims to learn the mean arm rewards. In our setting, it holds when the assumed underlying distribution is single-parameter, or when only one of the parameters of a multi-parameter distribution is unknown. Alternatively, it can be interpreted as identifying and correcting distribution shifts by updating a reference point in the distribution (e.g., adjusting the mean). For instance, a bank may want to adjust for increases in average credit scores \citep{avg-CS} over time. More specifically, we will let  $\hat{\parameter}^y_{g,t}$ be the $\tau$-th percentile of $\hat{f}^y_{g,t}(x)$. We discuss potential limitations of Assumption~\ref{as:distributions} in Section~\ref{sec:conclusion}, and present an extension to a case with two unknown parameters in Online Appendix~\ref{app:two_parameters}.

Under Assumption~\ref{as:distributions}, data bias can be quantified as the mistmatch between the estimated $\hat{\parameter}^y_{g,t}$ and true parameter ${\parameter}^y_{g}$. In particular, we use the mean absolute error $\mathbb{E}[|\hat{\parameter}^y_{g,t}-{\parameter}^y_{g}|]$ to measure the bias, where the randomness is due to $\hat{\parameter}^y_{g,t}$, the estimate of the unknown parameter based on data collected up to time $t$.

\textbf{Algorithm choice without debiasing.} 
Let ${\alpha}^y_{g}$ be the fraction of group $g$ agents with label $y$. A loss-minimizing fair algorithm selects its thresholds $\theta_{g,t}$ at time $t$ as follows: 
\begin{align}
\min_{\theta_{a,t}, \theta_{b,t}} &~~ \sum_{g\in \{a,b\}} \alpha^1_{g}\int_{-\infty}^{\theta_{g,t}} \hat{f}^1_{g,t}\mathrm{d}x + \alpha^0_{g}\int_{\theta_{g,t}}^{\infty} \hat{f}^0_{g,t}\mathrm{d}x  \hspace{0.2in}\text{s.t. } ~~ \mathcal{C}(\theta_{a,t}, \theta_{b,t}) = 0~.
    \label{eq:alg-obj}
\end{align}
Here, the objective is the misclassification error, and $\mathcal{C}(\theta_a, \theta_b) = 0$ is the fairness constraint imposed by the firm, if any. For instance, $\mathcal{C}(\theta_{a,t},  \theta_{b,t}) = \theta_{a,t} - \theta_{b,t}$ {would impose the \emph{same decision rule} constraint, or  $\mathcal{C}(\theta_{a,t}, \theta_{b,t}) = \int_{\theta_{a,t}}^\infty 
\hat{f}^1_{a,t}(x)\mathrm{d}x - \int_{\theta_{b,t}}^\infty \hat{f}^1_{b,t}(x)\mathrm{d}x$ would restrict the decision space to thresholds that meet the \emph{equality of opportunity} (true positive rate parity) constraint.} Note that the objective function and the fairness constraint are both affected by any inaccuracies in the current estimates $\hat{f}^y_{g,t}$. That is why a biased training dataset can lead to both loss of accuracy and loss in desired fairness.
\section{\texttt{Active Debiasing} Algorithm with Bounded Exploration} 
\label{sec:single-group}

In this section, we present the \texttt{active debiasing} algorithm, using both \emph{exploitation} (following the currently optimal decision rules of \eqref{eq:alg-obj}) and \emph{exploration} (allowing deviations up to a \emph{lowerbound} $LB_t$) to remove biases from the estimates $\hat{f}^y_{g,t}$. Although the deviations may lead to admitting some unqualified (label 0) agents, they help reduce biases in $\hat{f}^y_{g,t}$, enhancing both classification accuracy and fairness. In this section, we drop the subscripts $g$ from the notation; when there are multiple protected groups, our algorithm can be applied to each group's estimates separately.

As noted in Section~\ref{sec:intro}, our algorithm will differ from prior works mainly as it introduces \emph{bounded exploration}: it includes a lower bound $LB_t$,
which captures the extent to which the decision maker is willing to deviate from the current classifier $\theta_t$, based on its current estimate $\hat{f}^0_{t}$. Formally,

\begin{definition}\label{def:UB-LB}
At time $t$, the firm selects a $\text{LB}_t$ such that $\text{LB}_t = (\hat{F}^{0}_t)^{-1}(2\hat{F}^{0}_t(\hat{\med}^0_t)-\hat{F}^{0}_t(\theta_t))$, where $\theta_t$ is the (current) threshold determined from \eqref{eq:alg-obj}, $\hat{F}_t^0$, $(\hat{F}^{0}_t)^{-1}$ are the cumulative distribution function (CDF) and inverse CDF of the estimates $\hat{f}_{t}^0$, respectively, and $\hat{\parameter}^{0}_t$ is (wlog) the $\tau$-th percentile of $\hat{f}_{t}^0$. 
\end{definition}

In more detail, we choose $LB_t$ such that $\hat{F}^0_t(\hat{\omega}^0_t)-\hat{F}^0_t(LB_t)=\hat{F}^0_t(\theta_t)-\hat{F}^0_t(\hat{\omega}^0_t)$; that is, such that $\hat{\omega}^0_t$ is the median in the interval $(LB_t, \theta_t)$ based on the current estimate of the distribution $\hat{F}^0_t$ at the beginning of time $t$. Then, once a new batch of data is collected, we update $\hat{\omega}^0_t$ to $\hat{\omega}^0_{t+1}$, the \emph{realized} median of the distribution between $(LB_t, \theta_t)$ based on the data observed during $[t, t+1)$. Once the underlying distribution is correctly estimated, (in expectation) we will observe the same number of samples between $(LB_t, \omega^0_t)$ and between $(\omega^0_t, \theta_t)$, and hence $\omega^0_t$ will no longer change. We also note that by selecting a high $\tau$-th percentile in the above definition, $\text{LB}_t$ can be increased so as to limit the depth of exploration. As shown in Theorem~\ref{thm:debiasing}, and in our numerical experiments, these thresholding choice will enable debiasing of the distribution estimates while controlling its costs.

Our \texttt{active debiasing} algorithm is described below. The pseudo-code is shown in Algorithm~\ref{alg:two}. 
\SetKwComment{Comment}{/* }{ */}
\RestyleAlgo{ruled}
\begin{algorithm}
\caption{\texttt{Active Debiasing} Algorithm with Bounded Exploration}\label{alg:two}
\textbf{Input}: fairness constraint $\mathcal{C}(\theta_{a,t=0}, \theta_{b,t=0})$, Sample size $N$, Batch size $S$, Percentile $\tau^y_g$\\
\KwResult{Fair classifier $\theta_{g,t}$}
$t \gets 0, \epsilon_{g,t=0} \gets 1$\\
$\hat{\med}^y_{g,t=0} \gets (\hat{F}^{y}_{g,t=0})^{-1}(\tau^y_g)$ for $y \in \{0,1\}$, $LB_{g, t=0} \gets (\hat{F}^{0}_{g,t=0})^{-1}(2\hat{F}^{0}_{g,t=0}(\hat{\med}^0_{g,t=0})-\hat{F}^{0}_{g,t=0}(\theta_{g,t=0}))$\\
\While{$i \leq N$}{ 
    $\text{Data\_trun}^y_g$ = [ ], $\text{Data}^y_g$ = [ ], $k \gets 0$, $\text{portion\_left}^y_g$ = $\hat{F}^y_{g,t}(LB_{t} \leq x \leq \hat{\med}^y_g)/\hat{F}^y_{g,t}(LB_{t} \leq x)$\\
  \While{$k \leq S$ and $i \leq N$}{
    \For{$g \in G = \{a,b\}$}{
        \uIf{$\theta_{g,t} \leq x$, or $LB_{g,t} \leq x \leq \theta_{g,t}  \text{ and } rand() \leq \epsilon_{g, t}$}
            {$\text{Decision} \gets 1$ (accept)}
        \Else
        {$\text{Decision} \gets 0$ (reject)}
        \textbf{Add} $x$ into $\text{Data}^y_g$ if accepted, into $\text{Data\_trun}^y_g$ with $\epsilon_{g,t}$ if $x \in (LB_{g, t}, \theta_{g,t})$ or $(\theta_{g,t},\infty)$
    }
  $i \gets i + 1, k = \min(\text{len}(\text{Data\_trun}^y_a),\text{len}( \text{Data\_trun}^y_b))$
  }
  $t \gets t + 1$\\
  $\hat{\med}^y_{g,t} \gets quantile(\text{Data\_trun}^y_g, \text{portion\_left}^y_g)$  \Comment*[r]{Update reference value using all collected samples from the batch}
  \textbf{Map} back from $\hat{\med}^y_{g,t}$ to the single unknown parameter in the estimated distribution $\hat{f}^y_{g,t}$\\
  \textbf{Retrain} the classifier, output new threshold $\theta_{g,t}$ and $LB_{g,t}$ \Comment*[r]{Update classifier using all collected samples so far}
  $\textbf{Update } \epsilon_{g,t}$  \Comment*[r]{Can be fixed schedule reduction or adaptive}
}
\end{algorithm}

{\begin{algorithms}[\texttt{Active debiasing}]\label{def:db-alg} At each time $t$, the algorithm proceeds as follows:

\textbf{Stage $\emptyset$: Find the algorithm's parameters.} Use the current distribution estimates $\hat{f}^y_{t}$ (parameterized by $\hat{\parameter}^{y}_t$) to find the loss-minimizing decision threshold $\theta_t$ by solving \eqref{eq:alg-obj}. Find the current lowerbound $\text{LB}_t$ from Definition~\ref{def:UB-LB}. Let $\epsilon_t$ be the current exploration probability (selected from a pre-determined sequence). 

\textbf{Stage I: Admit agents and collect data.} New agents $({x}^\dagger, {y}^\dagger)$ arrive during $[t, t+1)$. Admit all agents with $x^{\dagger}\geq \theta_t$ (this is ``exploitation''). Further,  if $\text{LB}_t\leq x^\dagger < \theta_t$, admit the agent with probability $\epsilon_t$ (this is ``bounded exploration''). The admitted agents in this stage constitute the new data for Stage II's updates. 
 
\textbf{Stage II: Update the distribution estimates based on the new data collected in Stage I.}
\begin{itemize} 
\setlength\itemsep{0.01in}
    \item \emph{Qualified agents' distribution update:} Identify new data with $\text{LB}_t \leq x^\dagger$ and ${y}^\dagger=1$. Use all such $x^\dagger$ with $\text{LB}_t \leq x^\dagger < \theta_t$, and such $x^\dagger$ with $\theta_t \leq x^\dagger$ with probability $\epsilon_t$, to update $\hat{\parameter}_t^1$ to $\hat{\parameter}_{t+1}^1$. (For example, when the reference point $\hat{\parameter}^{y}_t$ is set to the median (the 50-th percentile), the parameter can be adjusted so that half the label 1 new data collected in Stage I will lie on each side of $\hat{\parameter}^y_{t+1}$.) 
    \item \emph{Unqualified agents' distribution update:} Identify new data with $\text{LB}_t \leq x^\dagger$ and ${y}^\dagger=0$. Use all such $x^\dagger$ with $\text{LB}_t \leq x^\dagger < \theta_t$, and such $x^\dagger$ with $\theta_t \leq x^\dagger$ with probability $\epsilon_t$, to update  $\hat{\parameter}_t^0$ to $\hat{\parameter}_{t+1}^0$. 
\end{itemize}
\end{algorithms}
}

\section{Theoretical Analysis of the \texttt{Active Debiasing} algorithm}\label{sec:analytical}

In this section, we analytically show that our proposed algorithm can recover unbiased estimates of the unknown parameter $\parameter^y$ in unimodal feature distributions (Theorem~\ref{thm:debiasing}). We then provide an error bound (on the number of wrong decisions) for our algorithm (Theorem~\ref{thm:regret}). We also highlight the impacts of using our algorithm in conjunction with existing algorithmic fairness interventions (Proposition~\ref{prop:fairness-debiasing}).

\subsection{Debiasing distribution estimates using the \texttt{active debiasing} algorithm} 

Before discussing the properties of \texttt{active debiasing}, we begin by analyzing the ability of two other algorithms in debiasing distribution estimates: \texttt{exploitation-only} (which only accepts agents with $x\geq \theta_t$, and uses no exploration or thresholding) and \texttt{pure exploration} (which accepts arriving agents at time $t$ who have $x<\theta_t$ with probability $\epsilon_t$, without setting any lower bound). The motivation for the choice of these two baselines is as follows: 

For the \texttt{exploitation-only} baseline, this algorithm represents a scenario where the decision maker is unaware of underlying data biases and makes no effort to address them. For example, in the context of loan applications, decision-makers select a decision threshold based on data from past decades to guide future decisions. However, this data may fail to reflect current trends, such as the upward trend in FICO credit scores over time \citep{avg-CS, avg-CS-2}. By relying on outdated information, the algorithm employs the decision threshold as-is and makes decisions accordingly. 
 
The \texttt{pure exploration} baseline on the other hand is inspired by the Bandit learning literature, and is also akin to debiasing algorithms proposed in recent work as discussed in Section~\ref{sec:related}. In this approach, decision-makers are aware of data biases and actively explore a broader range of samples to mitigate these biases. However, this baseline does not account for the costs associated with exploration—an important factor in high-stakes decision-making. Since the cost of accepting an unqualified applicant can often far exceed the cost of rejecting a qualified one, this omission makes the \texttt{pure exploration} approach less practical in scenarios where exploration costs must be carefully managed.

\begin{theorem}\label{thm:exploit-only_and_pure-exploration}
The \texttt{exploitation-only} algorithm overestimates $\parameter^y$, i.e., $\lim_{t\rightarrow \infty} \mathbb{E}[\hat{\parameter}^y_t]>\parameter^y, \forall y$. The \texttt{pure exploration} algorithm recovers unbiased estimates of $\parameter^y$, i.e., $\hat{\parameter}^y_{t} \xrightarrow{{a.s.}} {\parameter}^y$ as $t\rightarrow \infty$, $\forall y$. 
\end{theorem}

The detailed proof is given in Online Appendix~\ref{thm:exploit-only_and_pure-exploration}. For \texttt{exploitation-only}, it proceeds by identifying a martingale sequences in the feature change of the observed agents, and applying the Azuma-Hoeffding inequality to obtain a bound on it. For \texttt{pure exploration}, it invokes the strong law of large numbers.

Theorem~\ref{thm:exploit-only_and_pure-exploration} shows that the ignoring censored feedback (as done by \texttt{exploitation-only}) will ultimately result in overestimation of the underlying distributions, but that conducting exploration (as done by \texttt{pure exploration}) can mitigate this bias in the long-run. That said, \texttt{pure exploration}'s debiasing comes at the expense of accepting agents with \emph{any} $x<\theta_t$ (incurring high exploration cost as further illustrated in Section~\ref{sec:simulations}). Theorem~\ref{thm:debiasing} shows that our proposed exploration and thresholding procedure in the \texttt{active debiasing} algorithm, which limits the depth of exploration to $\text{LB}_t<x<\theta_t$, can still recover unbiased estimates of underlying unimodal feature distributions.

\begin{theorem}\label{thm:debiasing}
Let ${f}^y$ and $\hat{f}^y_{t}$ be the true feature distribution and their estimates at time $t$, with respective $\tau$-th percentiles $\parameter^y$ and $\hat{\parameter}_t^y$. Assume these are unimodel distributions, $\epsilon_t>0, \forall t$, and {$\hat{\parameter}_t^0\leq\theta_t\leq \hat{\parameter}_t^1, \forall t$}. Then, using the \texttt{active debiasing} algorithm, \textbf{(a)} if $\hat{\parameter}^y_{t}$ is underestimated (resp. overestimated), then  $\mathbb{E}[\hat{\parameter}^y_{t+1}] \geq \hat{\parameter}_t^y,$ (resp. $\mathbb{E}[\hat{\parameter}^y_{t+1}] \leq \hat{\parameter}_t^y$) $\forall t, \forall y$. \textbf{(b)} $\{\hat{\parameter}^y_{t}\}$ converges to ${\parameter}^y$ {almost surely} as $t\rightarrow \infty$, $\forall y$. 
\end{theorem}

\begin{proofsketch}
We provide a proof sketch for debiasing $\hat{f}_t^0$ which highlights the main technical challenges addressed in our analysis. The detailed proof is given in Online Appendix~\ref{thm:debiasing}. 
Our proof involves the analysis of statistical estimates $\hat{\parameter}^0_{t}$ based on data collected from \emph{truncated} distributions. In particular, by bounding exploration, our algorithm will only collect data with features $x\geq \text{LB}_t$, and can use only this truncated data to build estimates of the unknown parameter of the distributions. 

Part \textbf{(a)} establishes that the sequence of $\{\hat{\parameter}^y_t\}$ produced by our \texttt{active debiasing} algorithm ``moves'' in the right direction over time. The main challenge in this analysis is that as the exploration and update intervals $[\text{LB}_t, \infty)$ are themselves adaptive, there is no guarantee on the number of samples in each interval, and therefore we need to analyze the estimates in finite sample regimes.
To proceed with the analysis, we assume the feature distribution estimates follow {unimodel} distributions (such as Gaussian, Beta, and the family of $alpha$-stable distributions) with $\parameter^0$ as reference points. We then consider the expected parameter update following the arrival of a batch of agents. Denote the current left portion in $(\text{LB}_t, \hat{\parameter}^0_t)$ as $p_1 := \frac{\hat{F}^0(\hat{\parameter}^0_t) -\hat{F}^0(\text{LB}_t)}{\hat{F}^0(\theta_t)-\hat{F}^0(\text{LB}_t)}$. Based on Definition ~\ref{def:UB-LB}, we can also obtain the current portion in $(\hat{\parameter}^0_t, \theta_t)$ denoted as $p_2 := \frac{\hat{F}^0(\theta_t) - \hat{F}^0(\hat{\parameter}^0_t)}{\hat{F}^0(\theta_t)-\hat{F}^0(\text{LB}_t)} = p_1$. The new expected estimates $\mathbb{E}[\hat{\parameter}^0_{t+1}]$ is the sample median in $(\text{LB}_t, \theta_t)$, where samples come from the true distribution. We establish that this expected update will be higher/lower than $\parameter^0_{t}$ if the current estimate is an under/over estimate of the true parameter. 

Then, in Part \textbf{(b)} we first show that the sequence of over- and under-estimation errors in $\{\hat{\parameter}^y_t\}$ relative to the true parameter ${\parameter}^y$ are supermartingales. By the Doobs Convergence theorem and using results from part \textbf{(a)}, these will converge to zero mean random variables with variance going to zero as the number of samples increases. This establishes that $\{\hat{\parameter}^y_t\}$ converges. It remains to show that this convergence point is the true parameter of the distribution. To do so, as detailed in the proof, we note that the density function of the sample median estimated on label 0 data collected in $[\text{LB}_t, \theta_t]$ is
\begin{align*}
\mathbb P(\hat{\parameter}^0_{t}=\nu) \mathrm{d}\nu= \frac{(2m+1)!}{m!m!}(\tfrac{F^0(\nu)-F^0(\text{LB}_t)}{F^0(\theta_t)-F^0(\text{LB}_t)})^m(\tfrac{F^0(\theta_t)-F^0(\nu)}{F^0(\theta_t)-F^0(\text{LB}_t)})^m \tfrac{f^0(\nu)}{F^0(\theta_t)-F^0(\text{LB}_t)}\mathrm{d}\nu
\end{align*}
which is a beta distribution {pushed forward} by $H(\nu):=\frac{F^0(\nu)-F^0(\text{LB}_t)}{F^0(\theta_t)-F^0(\text{LB}_t)}$; this is the CDF of the truncated $F^0$ distribution in $[\text{LB}_t, \theta_t]$. We then establish that the convergence point will be the true median of the underlying distribution. 
\end{proofsketch}

\subsection{Error bound analysis}\label{sec:regret-analysis}

We next compare errors (measured as the number of wrong decisions) of our \texttt{adaptive debiasing} algorithm against those made by an oracle with knowledge of the true underlying distributions; {this provides an assessment of \emph{decision making} costs incurred to overcome data biases}. We measure the performance using 0-1 loss, $\ell(\hat{y_i},y_i)=\mathbbm{1}[\hat{y_i}\neq y_i]$, where $\hat{y}_i$ and $y_i$ denote the predicted and true label of agent $i$, respectively. We consider the error accumulated when updating the estimates using a total of $m$ batches of data. We split the total $T$ samples that have arrived during $[t, t+1)$ into four groups, corresponding to four different distributions ${f}^y_g$. Specifically, we use ${b}^y_{g,t}$ to denote the number of samples from each label-group pair at round $t \in \{0, \ldots, m\}$. We update the unknown distribution estimates once all batches meet a size requirement $s$, i.e, once $\min({b}^y_{g,t}) \geq s, \forall y, \forall g$. The error of our algorithm is given by:
\begin{align*}
    &\text{Error} = {\mathbb{E}}[Error_{Adaptive} - Error_{Oracle}] \\
    & = \sum_{t} \sum_{i=1}^{{b}^0_{a,t}+{b}^1_{a,t}+{b}^0_{b,t}+{b}^1_{b,t}} \mathop{\mathbb{E}}_{(x_i,y_i,g_i)\sim D}\Big[\ell(h_{\theta_{t,g}}(x_i,g_i),y_i)\Big] - \sum_{i=1}^{T} \displaystyle \mathop{\mathbb{E}}_{(x_i,y_i,g_i)\sim D}\Big[\ell(h^*_{\theta_g}(x_i,g_i),y_i)\Big] 
\end{align*}

Notice that, the error terms, $Error_{Adaptive}$ and $Error_{Oracle}$, share the same loss expression in Eq.~\ref{eq:alg-obj}. The differences lie in the choice of the decision threshold. $Error_{Adaptive}$ is calculated with the current (potentially) biased classifier $\theta_{g,t}$, whereas $Error_{Oracle}$ is calculated with the optimal decision threshold $\theta_g^*$. The following theorem provides an upper bound on the error incurred by \texttt{active debiasing}. 

\begin{theorem}~\label{thm:regret}
Let $\hat{f}^y_{g,t}(x)$ be the estimated feature-label distributions at round $t \in \{0, \ldots, m\}$. We consider the threshold-based, group-specific, binary classifier $h_{\theta_{g,t}}$, and denote the Rademacher complexity of the classifier family $\mathcal{H}$ with $n$ training samples by $\mathcal{R}_n(\mathcal{H})$. Let $\theta_{g,t}$ be a $v$-approximately optimal classifier based on data collected up to time $t$. At round $t$, let $N_{g,t}$ be the number of exploration errors incurred by our algorithm, $n_{g,t}$ be the sample size at time $t$ from group $g$, $d_{\mathcal{H}\Delta\mathcal{H}}(\tilde{D}_{g,t}, D_g)$ be the distance between the true unbiased data distribution $D_g$ and the current biased estimate $\tilde{D}_{g,t}$, and $c(\tilde{D}_{g,t}, D_g)$ be the minimum error on an algorithm trained on unbiased and biased data. Then, with probability at least $1 - 4\delta$ with $\delta >0$, the active debiasing algorithm's error is bounded by:
{\small \[
\text{Err.} \leq \sum_{g,t} \Big[ \underbrace{2v}_{\text{$v$-approx.}} + \underbrace{4\mathcal{R}_{n_{g,t}}(\mathcal{H}) + \tfrac{4}{\sqrt{n_{g,t}}} + \sqrt{\tfrac{2\ln(2/\delta)}{n_{g,t}}}}_{\text{empirical estimation errors}} + \underbrace{N_{g,t}}_{\text{explor.}} + \underbrace{d_{\mathcal{H}\Delta\mathcal{H}}(\tilde{D}_{g,t}, D_g)+2c(\tilde{D}_{g,t}, D_g)}_{\text{source-target distribution mismatch}} \Big]\]}
\end{theorem}

\begin{proofsketch}
We provide a description of the steps involved in finding the above error bound. More details on the definitions of the distance measure $d_{\mathcal{H}\Delta\mathcal{H}}$, the error term $c(\cdot)$, and the exploration error term $N_{g,t}$, along with a a detailed proof, are given in Online Appendix~\ref{app:regret_proof}.

This proof is based on a reduction from fair-classification to a sequence of cost-sensitive classification problems, as proposed and also used to obtain error bounds in \citep{agarwal2018reductions}, and in learning under source and target distribution mismatches as proposed in \citep{ben2010theory}.  We adapt these to our bounded exploration setting. In order to find our algorithm's error bound, we proceed through five steps. The first step is to view each individual update of the fair threshold classifier as a saddle point problem, which can be solved efficiently by the exponentiated gradient reduction method introduced in \citep{agarwal2018reductions}. Therefore, we have an \emph{approximately} optimal classifier with suboptimality level $v$ (Step 1). Second, based on the solution output from the reduction method, we find the empirical estimation error based on data from the biased distributions (Step 2). Thirdly, using results from \citep{ben2010theory}, we bound the error on the target (unbiased) distribution when the algorithm is obtained from the biased source domain (Step 3). Then, we will evaluate the impact of exploration errors made by our debiasing algorithm (Step 4). Finally, we aggregate over $t$ rounds of updates (Step 5).  
\end{proofsketch}

From the expression in Theorem~\ref{thm:regret}, we can see that the errors (wrong decisions made in order to, and while, overcoming data biases and censored feedback) incurred by our algorithm consist of four types: errors due to approximation of the optimal (fair) classifier at each round, empirical estimation errors, exploration errors, and errors due to our biased training data (viewed as source-target distribution mismatches); the latter two are specific to our \texttt{active debiasing} algorithm. In particular, as we collect more samples, $n_{g,t}$ will increase. Hence, the empirical estimation errors decrease over time. Moreover, as the mismatch between $\tilde{D}_{g,t}$ and $D_g$ decreases using our algorithm (by Theorem~\ref{thm:debiasing}), the error due to target domain and source domain mismatches also decrease. In the meantime, our exploration probability $\epsilon_t$ also becomes smaller over time, decreasing the exploration error term $N_{g,t}$. 

\subsection{\texttt{Active debiasing} and fairness {interventions}}\label{sec:two-group}

Lastly, we consider the impacts of imposing fairness constraints (the constraints in \eqref{eq:alg-obj}), which will lead to a change in the selected classifiers, together with our proposed debiasing algorithm. Let $\theta^{F}_{g,t}$ and $\theta^{U}_{g,t}$ denote the fairness constrained and unconstrained decision rules obtained from \eqref{eq:alg-obj} at time $t$ for group $g$, respectively. We say group $g$ is being over-selected (resp. under-selected) following the introduction of fairness constraints if $\theta^{F}_{g,t}<\theta^{U}_{g,t}$ (resp. $\theta^{F}_{g,t}>\theta^{U}_{g,t}$). Below, we show how such over/under-selections can differently affect the debiasing of estimates on different agents. 

In particular, let the speed of debiasing be the rate at which  $\mathbb{E}[|\hat{\parameter}^y_t-\parameter^y|]$ decreases with respect to $t$; then, for a given $t$, an algorithm for which this error is larger has a slower speed of debiasing. The following proposition identifies the impacts of different fairness constraints on the speed of debiasing attained by our \texttt{active debiasing} algorithm. 

\begin{proposition}\label{prop:fairness-debiasing}
Let  ${f}^y_g$ and $\hat{f}^y_{g,t}$ be the true and estimated feature distributions, with respective medians $\parameter^y$ and $\hat{\parameter}_t^y$. Assume these are unimodel distributions, and \texttt{active debiasing} is applied. If group $g$ is over-selected (resp. under-selected), i.e.,  $\theta^{F}_{g,t}<\theta^{U}_{g,t}$ (resp. $\theta^{F}_{g,t}>\theta^{U}_{g,t}$), the speed of debiasing on the estimates $\hat{f}^y_{g,t}$ will decrease (resp. increase).  
\end{proposition}

The proof appears in Online Appendix~\ref{app:proof_fairness_debiasing}. Proposition~\ref{prop:fairness-debiasing} highlights the following implications of using both fairness rules and our active debiasing efforts. Some fairness constraints (such as equality of opportunity) can lead to an increase in opportunities for (here, over-selection of) agents from disadvantaged groups, while others (such as same decision rule) can lead to under-selection from that group. Proposition~\ref{prop:fairness-debiasing} shows that \texttt{active debiasing} may in turn become faster or slower at debiasing estimates on this group. 

Intuitively, over-selection provides increased opportunities to agents from a group (compared to an unconstrained classifier). In fact, the reduction of the decision threshold to $\theta^{F}_{g,t}$ can itself be interpreted as introducing exploration (which is separate from that introduced by our debiasing algorithm). When a group is over-selected under a fairness constraint, the fairness-constrained threshold $\theta^F_{g,t}$ will be lower than the unconstrained threshold $\theta^U_{g,t}$. Therefore, the exploration range will be narrower, which means by adding a fairness constraint, the algorithm needs to wait and collect more samples (takes a longer time) before it manages to collect sufficient data to accurately update the unknown distribution parameter, and hence, it has a slower debiasing speed. More broadly, these findings contribute to our understanding of how fairness constraints can have long-term implications beyond the commonly studied fairness-accuracy tradeoff when we consider their impacts on data collection and debiasing efforts.

{\section{Noisy Exploration: Introduction of an Intermediate Exploration Action}\label{sec:mdp} 
We next propose to extend Algorithm~\ref{def:db-alg} by incorporating additional \emph{intermediate exploration} actions which can provide noisy information about unobserved labels at a lower cost. Specifically, we assume the decision maker can offer an intermediate option within the exploration range (e.g., a small loan, an internship). Let the probability of an explored unqualified agent ($y=0$) fulfilling the requirements of this intermediate action successfully be $\gamma$. If an agent fails to fulfill these requirements, their true qualification state ($y=0$) will be correctly identified by the decision maker. Otherwise, the decision maker mistakenly labels this agent as $y=1$, reflecting the noisy nature of the intermediate action. Therefore, compared to binary exploration (considered earlier), samples labeled 1 via intermediate actions are cheaper to obtain but unreliable, potentially slowing debiasing. To formally highlight this trade-off between the debiasing speed and the costs incurred from exploration, we consider the problem of making intermediate vs. uniform exploration decisions in a two-stage MDP framework.

In this two-stage MDP, the decision maker faces a classification problem at both times $t=1$ and $t=2$, and its action set is its exploration choice $A=\{I, U, N\}$, denoting Intermediate (noisy) exploration, Uniform (accurate) exploration, or No exploration, respectively. As exploration (either action $I$ or $U$) is costly, the optimal policy at the terminal stage $t=2$ is to not explore ($a_2=N$) and stick with the current (potentially biased) loss-minimizing classifier. However, the decision maker can choose to explore at $t=1$ to mitigate data biases, which could help improve decision accuracy at time step $t=2$. Therefore, we use this MDP framework to assess the decision maker's choice between intermediate vs. uniform exploration at $t=1$.

The expected cumulative cost for the two-stage MDP is given by:
\begin{align}
    \mathbb{E}[\mathcal{L}(a_1, \hat{f}_1^0, \hat{f}_1^1)] =  \mathbb{E}[\mathcal{L}^{\text{exp-cost}}_1(a_1, \hat{f}_1^0, \hat{f}_1^1)] + \sum_{t \in \{1,2\}} \mathbb{E}[\mathcal{L}^{\text{miss-cost}}_t(\hat{f}_t^0, \hat{f}_t^1)]
\end{align}
where $\hat{f}^y_t$ denote the estimated underlying distributions at time $t$, $\mathcal{L}^{\text{exp-cost}}_1$ is the exploration cost incurred at time $t=1$, and $\mathcal{L}^{\text{miss-cost}}_t$ is the missclassification error cost at time $t$. (These are detailed shortly.) 

We begin by defining the cost for each type of decision error (i.e., true label not matching the prescribed accept/reject decision). (1) A \emph{missed opportunity cost} for rejecting qualified ($y=1$) agents, appearing at two levels: $L^h_1$ and $L^l_1$, with $L_1^l<L_1^h$. (For example, $L^h_1$ could reflect losing the future business of qualified loan applicants who are mistakenly rejected, whereas $L^l_1$ reflects the loss of business of qualified loan applicants who are not rejected, but rather receive a smaller loan than they requested.)
(2) A \emph{net loss} incurred for accepting unqualified ($y=0$) agents, appearing at two levels $L^h_2$ and $L_2^l$, with $L_2^l<L_2^h$. (For example, $L_2^h$ could capture the loss of principal on a large loan that an unqualified agent fails to pay off, whereas $L_2^l$ is the loss on a micro-loan that is not paid off.)

Let $N_t$ denote the number of agents arriving during $[t, t+1)$. Then, the expected misclassification costs at each time step $t$ can be expressed as follows:
\begin{align}
    \mathbb{E}[\mathcal{L}^{\text{miss-cost}}_t(\hat{f}_t^0, \hat{f}_t^1))] = N_t \Big \{L^h_1 \alpha^1 \int_{-\infty}^{\hat{\theta}_t} f^1(x)dx + L^h_2 \alpha^0 \int_{\hat{\theta}_t}^{\infty} f^0(x)dx\Big\}
\end{align}
Note that the difference in $L^{\text{miss-cost}}_t$ under $a_1=I$ vs. $a_1=U$ will emerge at $t=2$, as the exploration choice affects the errors in $\hat{f}^y_2$, and therefore leads to differences in the decision thresholds $\hat{\theta}^U_t$ and $\hat{\theta}^{I}_t$. 

The expected exploration costs, on the other hand, differ more. Specifically:
\begin{align}
    \mathbb{E}[\mathcal{L}^{\text{exp-cost}}_1(U, \hat{f}_1^0, \hat{f}_1^1)] &=  N_t \Big \{ -L^h_1 \epsilon_t\alpha^1 \int_{LB^U_t}^{\hat{\theta}^U_t} f^1(x)dx + L^h_2 \epsilon_t \alpha^0 \int_{LB^U_t}^{\hat{\theta}^U_t} f^0(x)dx\Big \}\notag\\
    \mathbb{E}[\mathcal{L}^{\text{exp-cost}}_1(I, \hat{f}_1^0, \hat{f}_1^1)] &=  N_t \Big \{ (-L_1^h + L_1^l) \epsilon_t\alpha^1 \int_{LB^{I}_t}^{\hat{\theta}^{I}_t} f^1(x)dx + L_2^l(1-\gamma) \epsilon_t \alpha^0 \int_{LB^{I}_t}^{\hat{\theta}^{I}_t} f^0(x)dx\Big \}
\end{align}
where $\epsilon_t$ denotes the exploration probability, while $\gamma$ denotes the probability of explored unqualified samples fulfilling the requirements of the intermediate action. 

We now compare the impact of making intermediate vs. uniform exploration decisions in this two-stage MDP. The first theorem shows that the noisy nature of intermediate actions can indeed slow down debiasing. Specifically, it shows that the second stage decision threshold is closer to the optimal threshold under uniform exploration than under intermediate actions.

\begin{theorem}~\label{thm:theta}
Consider the described two-stage MDP. Let $\theta^*$ be the optimal loss-minimizing classifier, and let $\hat{\theta}^{I}_{2}$ and $\hat{\theta}^U_{2}$ denote the (potentially suboptimal) loss-minimizing classifiers selected at $t=2$, given exploration decisions $a_1=I$ and $a_1=U$ at $t=1$, respectively. Assume $\hat{f}_t^y$ and $f^y$ (the estimated and true feature distributions) are Gaussian with equal variance. Then, $\mathbb{E}[|\theta^* - \hat{\theta}^U_2]] \leq \mathbb{E}[|\theta^* - \hat{\theta}^{I}_2|]$.  
\end{theorem}

The main challenge in proving this theorem is to evaluate the impact of the explored data quality on mitigating the data bias, measured as the mean absolute error $\mathbb{E}[|\hat{\parameter}^y_{t}-{\parameter}^y|]$. With the Gaussian distribution assumption, we can analytically solve for the loss-minimizing decision threshold $\hat{\theta}$ in terms of $\hat{\parameter}^y_{t}$, enabling a comparison of debiasing speed. The detailed proof is given in Online Appendix~\ref{app:proof_mdp}.

The following theorem shows that despite the slow-down in debiasing speed, adopting intermediate exploration actions may be preferred as it lowers the cumulative loss.

\begin{theorem}~\label{thm:mdp}
Consider the described two-stage MDP. Assume $f^y$ are Gaussian distributions with the same variance, with $f^1$ having a larger mode that $f^0$. Let $\epsilon = 1$. If $(1-\frac{N_2}{N_1})(L^h_2\alpha^0-L^h_1\alpha^1) \geq L^l_2(1-\gamma)\alpha^0$, then $\mathbb{E}[\mathcal{L}(I, \hat{f}_1^0, \hat{f}_1^1)] \leq \mathbb{E}[\mathcal{L}(U, \hat{f}_1^0, \hat{f}_1^1)]$.
\end{theorem}

The detailed proof is given in Online Appendix~\ref{app:proof_mdp}. Intuitively, the condition $(1-\frac{N_2}{N_1})(L^h_2\alpha^0-L^h_1\alpha^1) \geq L^l_2(1-\gamma)\alpha^0$ in this theorem implies that a combination of the following conditions can make the intermediate action desirable: there are many unqualified individuals (high $\alpha^0$), there is considerably more loss from uniform exploration than from intermediate actions ($L_2^h$ considerably higher than $L_2^l$), the intermediate action is sufficiently accurate at identifying unqualified individuals (high $\gamma$), and/or there are not significantly more agents at time $t=2$ so that the loss of accuracy due to intermediate actions, as highlighted in Theorem~\ref{thm:theta}, is tolerable ($N_2$ is smaller, or not significantly larger, than $N_1$). Together, our findings highlight how the decision-maker should account for the trade-off between quickly reaching a more accurate loss-minimizing classifier, and the cumulative cost incurred to accomplish such debiasing. 
}
\section{Numerical Experiments}\label{sec:simulations}

We now illustrate the performance of our algorithms using both synthetic data and real-world \emph{Adult}~\citep{Dua:2019}, \emph{Retiring Adult}~\citep{ding2021retiring}, and the \emph{FICO} credit score~\citep{hardt2016equality} datasets. Our code is available at: \url{https://github.com/INFORMSJoC/2024.0651} \citep{Yang2025}. 

{The \emph{Adult} and \emph{Retiring Adult} datasets both include demographic information (e.g., age, gender, race, education, occupation, etc.) and are used to predict whether an individual earns more than \$50,000 annually. The \emph{Adult} dataset, sourced from the 1994 Census database, contains 48,842 samples, while the \emph{Retiring Adult} dataset is more recent and significantly larger, with 1,664,500 samples. Additionally, the \emph{FICO} dataset, containing 174,048 samples, provides credit score distribution information across different racial groups, with the objective of predicting whether an individual will default. 

Throughout, we either choose a \emph{fixed} schedule for reducing the exploration frequencies $\{\epsilon_t\}$, or reduce these \emph{adaptively} as a function of the estimated error. For the latter, we select a range (e.g., above the classifier for label 0/1) and adjust the exploration frequency proportional to the discrepancy between the number of observed classification errors in this interval relative to the number expected given the distribution estimates. 

\begin{figure}[ht]
	\centering
	\subfigure[Baselines]{
	    \includegraphics[width=0.23\textwidth]{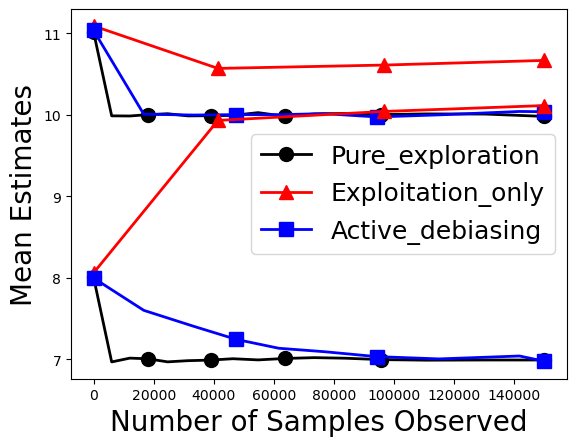}\label{fig:base_over}
	}
        \subfigure[Acc. performance]{
	    \includegraphics[width=0.23\textwidth]{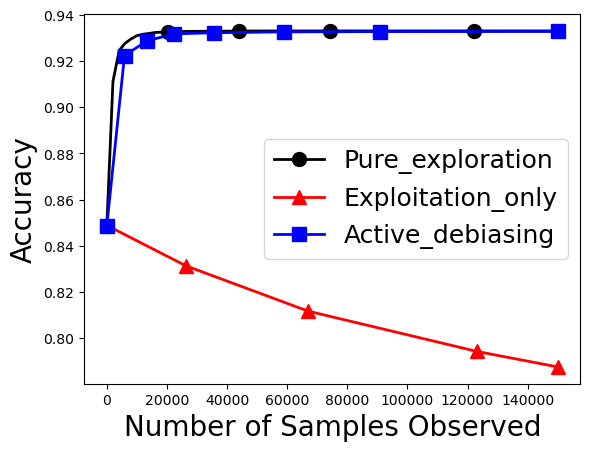}\label{fig:acc-performance}
	}
        \subfigure[Regret]{
	    \includegraphics[width=0.23\textwidth]{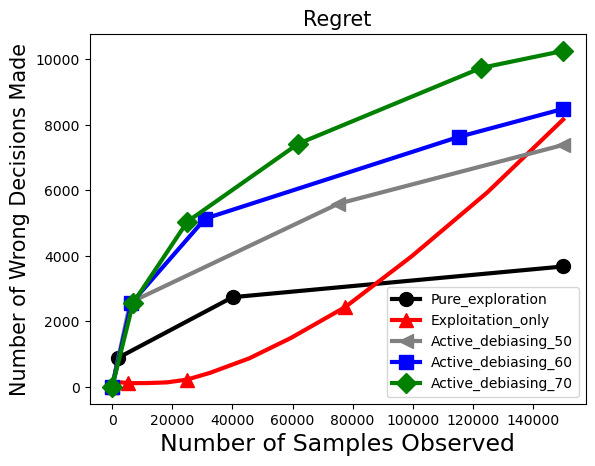}
		\label{fig:regret}
	}
	\subfigure[Weighted regret]{
	    \includegraphics[width=0.23\textwidth]{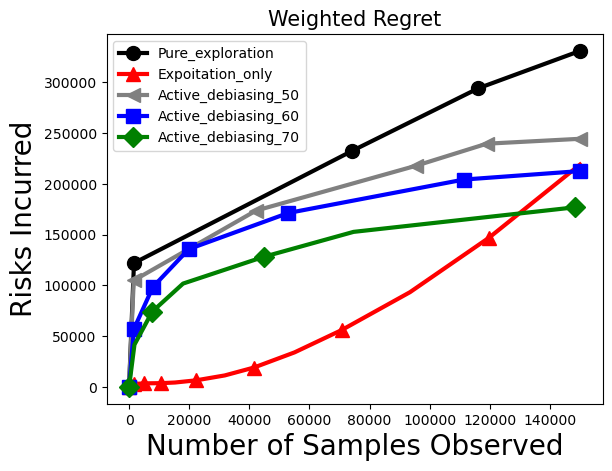}
		\label{fig:weighted-regret}
	}
	\caption{Debiasing performance, (weighted) regret, of \texttt{active debiasing} vs. baselines}\label{fig:debiasing_performance}
    \vspace{-0.2in}
\end{figure}

\textbf{Comparison with the \texttt{exploitation-only} and \texttt{pure exploration} baselines:} Our first experiments in Fig.~\ref{fig:debiasing_performance} compare our algorithm against two baselines on synthetic data. The underlying distributions are Gaussian; we let $f^1$ be overestimated with $\hat{\parameter}^1=11$ and true parameter $\parameter^1=10$  (parameter debiasing shown in the top lines), and $f^0$ be overestimated with $\hat{\parameter}^0=8$ and true parameter $\parameter^0=7$ as reference points. No fairness constraint is imposed. Our algorithm sets $\tau^1=50$ and $\tau^0=60$ percentiles, and exploration frequencies $\epsilon_t$ are selected adaptively by both our algorithm and \texttt{pure exploration}. The results in Fig.~\ref{fig:base_over}, comparing success in debiasing and debiasing speed, are consistent with Theorem~\ref{thm:exploit-only_and_pure-exploration}: \texttt{exploitation-only} overestimates the distributions due to adaptive sampling bias, and \texttt{pure exploration} and \texttt{active debiasing} both successfully debiases the estimates, with \texttt{pure exploration} debiasing \emph{faster}; this is because \texttt{pure exploration} observes samples with \emph{lower} features $x$ than \texttt{active debiasing}, and so can use these to reduce its estimate faster. In Fig.~\ref{fig:acc-performance}, we observe that, as expected, both the \texttt{pure exploration} and \texttt{active debiasing} algorithms improve accuracy, while the \texttt{exploitation-only} algorithm does not.

\textbf{Regret and Weighted Regret:} Figs.~\ref{fig:regret} and~\ref{fig:weighted-regret} compare the regret and weighted regret of the algorithms. Regret is measured as the difference between the number of false-negative (FN) and false-positive (FP) decisions of an algorithm vs. the oracle loss-minimizing algorithm derived on unbiased data. Formally, regret is defined as in Section~\ref{sec:regret-analysis}; weighted regret is defined similarly, but also adds a weight to each FN or FP decision, with the weight exponential in the distance of the feature of the admitted agent from the classifier. We let $f^1$ and $f^0$ be overestimated with $\hat{\parameter}^1=9$ and $\hat{\parameter}^0=6$, and their true parameters $\parameter^1=10$ and $\parameter^0=7$, respectively. We observe that \texttt{exploitation-only}'s regret is super-linear, failing to debias and accumulating errors from overestimation. On the other hand, while algorithms that explore ``deeper'' have lower regret (\texttt{pure exploration} $<$ \texttt{active debiasing} with $\tau^0=50 <$ \texttt{active debiasing} with $\tau^0=60$ in Fig.~\ref{fig:regret}), they have higher weighted regret (the order is reversed in Fig.~\ref{fig:weighted-regret}). 

\textbf{Computational performance:} Lastly, Table~\ref{tab1} compares the runtime of our algorithm against the baseline algorithms. All algorithms are implemented in Jupyter Notebook 6.4.12 and run with Python 3.9.13 on a MacBook Pro with 1.4GHz Quad-Core Intel Core i5 processor and Intel Iris Plus Graphics 645 1536 MB. From Table~\ref{tab1}, we observe that the \texttt{exploitation-only} algorithm exhibits the fastest performance as it does not explore (although leading to overestimated distribution estimates as indicated in Fig.~\ref{fig:debiasing_performance}). 
The \texttt{pure exploration} algorithm requires 21.6\% more time compared to the \texttt{exploitation-only} algorithm as it explores samples to mitigate the data bias. Our \texttt{adaptive debiasing} algorithm takes slightly more time (6.8\%) compared to \texttt{pure exploration} due to additional steps involved in calculating the $LB$ for bounded exploration decisions. 

\begin{table}[ht]
\centering
\begin{small}
    \caption{Computational performance comparison. Unit: second.\label{tab1}}
    \begin{tabular}{|c|c|c|} 
        \hline
        Exploitation-only baseline & Pure exploration baseline & Adaptive debiasing algorithm\\
        \hline
        114.28 (+/-0.40) & 138.98 (+/-0.53) & 148.41 (+/-0.44)\\
        \hline
    \end{tabular}
\end{small}
\end{table}

\textbf{\texttt{Active debiasing} on real-world datasets:} Fig.~\ref{fig:adult} illustrates the performance of our algorithm on the \emph{Adult} dataset. Data is grouped based on race (White ${G}_a$ and non-White ${G}_b$), with labels $y=1$ for income $>\$50k/year$. 
A one-dimensional feature $x \in \mathbb{R}$ is constructed by conducting logistic regression on four quantitative and qualitative features (education number, sex, age, workclass), based on the initial training data. Using an input analyzer, we found Beta distributions as the best fit to the underlying distributions (details are provided in Online Appendix~\ref{sec:more-experiments}). We use {2.5\%} of the data to obtain a biased estimate of the parameter $\alpha$. The remaining data arrives sequentially. We use $\tau^1=50$ and $\tau^0=60$ and a fixed decreasing $\{\epsilon_t\}$, with the equality of opportunity fairness constraint imposed throughout. We observe that our proposed algorithm can debias estimates across groups and for both labels in the long run with sufficient samples: in \emph{Adult}, as there are only {1080} samples for label 1 agents from $G_b$, the final estimate differs from the true value. Fig.~\ref{fig:group-b-extra-data} verifies that this estimate would have been debiased in the long run with additional (synthetically generated) samples from the underlying population. {Furthermore, we can observe from Fig.~\ref{fig:adult_acc-fair_performance} that both accuracy and equality of opportunity fairness improve by using our proposed algorithm.}

We also conduct similar experiments on the \emph{FICO credit score} and \emph{Retiring Adult} datasets (details about preparing these datasets are given in Online Appendix~\ref{sec:more-experiments}). Fig.~\ref{fig:fico_large} and Fig.~\ref{fig:retiring_adult}  illustrate the performance of \texttt{active debiasing} on these datasets, respectively. We again observe that our algorithm is successful at debiasing distribution estimates on both groups and labels. Furthermore, it shows that mitigating data biases through our algorithm leads to improvements in both accuracy and equality of opportunity fairness of the classifiers learned on the debiased data.

\begin{figure}[ht]
	\centering
	\subfigure[Adult, Debiasing $G_a$]
	{
		\includegraphics[width=0.21\textwidth]{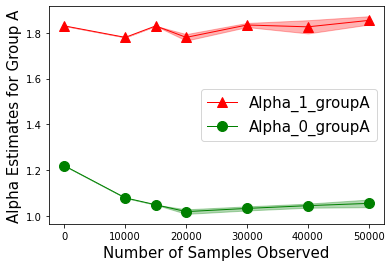}
		\label{fig:adult-white}
	}
	\subfigure[Adult, Debiasing $G_b$]
	{
		\includegraphics[width=0.21\textwidth]{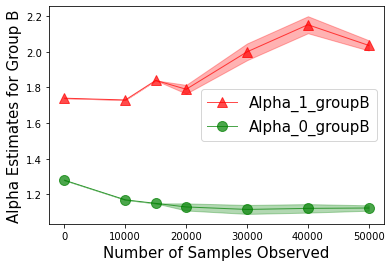}
		\label{fig:adult-nonwhite}
	}
	\subfigure[$G_b$ with more data]
	{
		\includegraphics[width=0.21\textwidth]{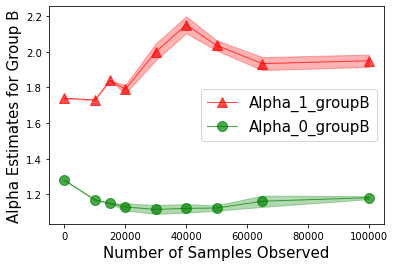}
		\label{fig:group-b-extra-data}
	}
         \subfigure[{Adult: Acc. - Fairness}]
	{
		\includegraphics[width=0.23\textwidth]{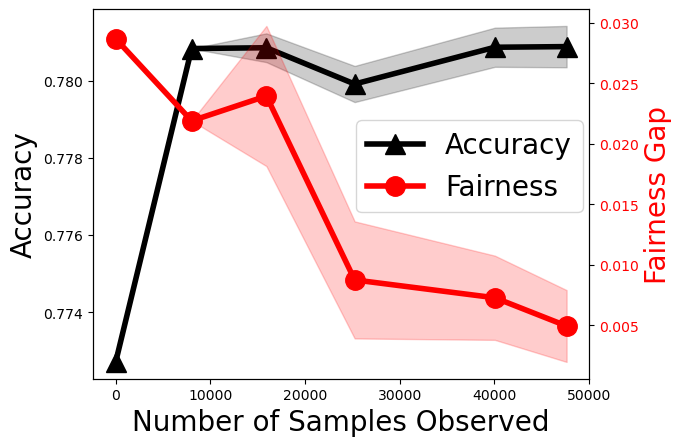}
		\label{fig:adult_acc-fair_performance}
	}
	\caption{\texttt{Active debiasing} on the \emph{Adult} dataset.}
	\label{fig:adult}
    \vspace{-0.2in}
\end{figure}

\begin{figure}[htbp!]
	\centering
	\subfigure[Active Debiasing on \emph{FICO}.]
	{
		\includegraphics[width=0.25\textwidth]{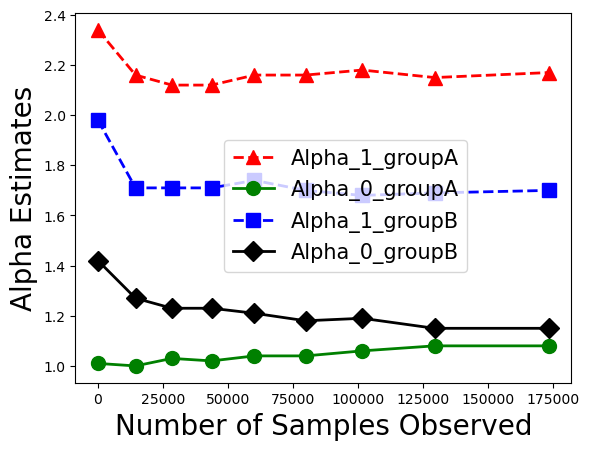}
		\label{fig:FICO_groupAandB_large}
	}
	\hspace{0.05in}
	\subfigure[Difference w.r.t. the true value.]
	{	\includegraphics[width=0.26\textwidth]{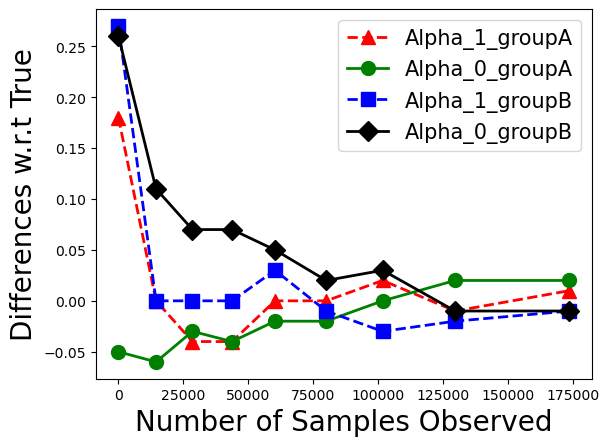}
		\label{fig:FICO_diff_large}
	}
   \hspace{0.05in}
	\subfigure[{\emph{FICO}: Acc. - Fairness}]
	{	\includegraphics[width=0.25\textwidth]{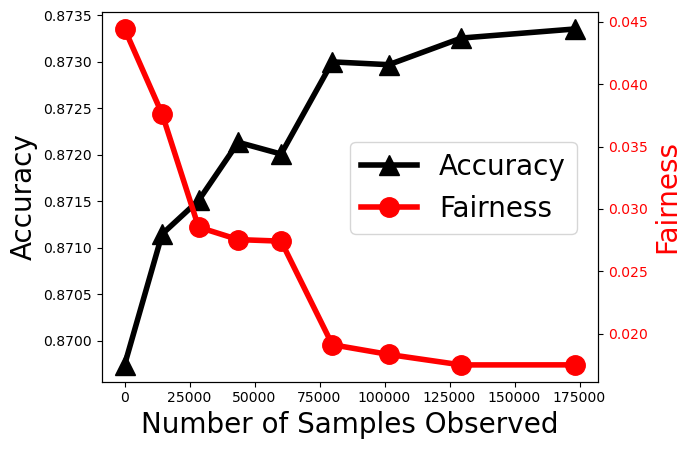}
		\label{fig:FICO_acc-fair_performance}
	}
	\caption{{\texttt{Active debiasing} on the \emph{FICO} dataset.}}
	\label{fig:fico_large}
\end{figure}

\begin{figure}[htbp!]
{
	\centering
	\subfigure[Debiasing on \emph{Retiring Adult}]
	{
    \includegraphics[width=0.25\textwidth]{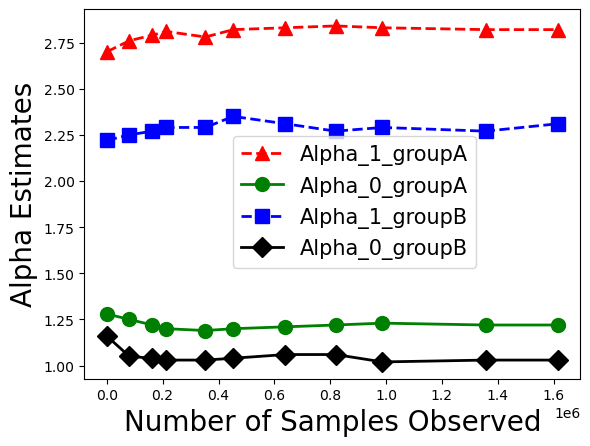}
		\label{fig:alpha_retiring}
	}
 \hspace{0.2in}
	\subfigure[Difference w.r.t. the true value]
	{
		\includegraphics[width=0.25\textwidth]{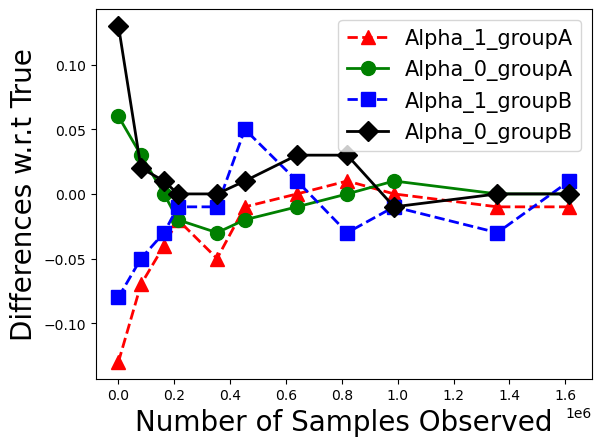}
		\label{fig:difference_retiring}
	}
	\subfigure[\emph{Retiring Adult}: Acc - Fairness]{
		    \includegraphics[width=0.28\textwidth]{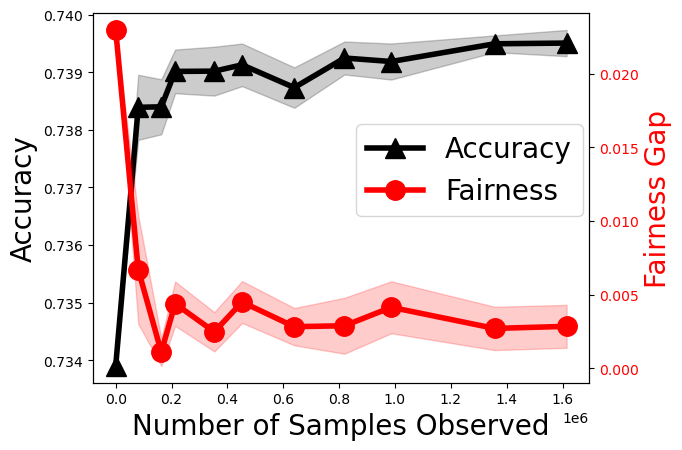}
		\label{fig:acc_fair_retiring}}
	\caption{{\texttt{Active debiasing} on the \emph{Retiring Adult} dataset.}}
	\label{fig:retiring_adult}}
\end{figure}

\textbf{Introduction of intermediate actions:} Fig.~\ref{fig:intermediate} compares the impacts of opting for uniform vs. intermediate actions on the accumulated costs and debiasing speed in our two-stage MDP framework. The underlying distributions are Gaussian. We let $f^1$ is underestimated with $\hat{\parameter}^1=9$ and $\parameter^1=10$, and $f^0$ is underestimated with $\hat{\parameter}^0=6$ and $\parameter^0=7$. We let $\gamma = 0.5$. From Fig.~\ref{fig:intermediate_performance}, we can see that the debiasing speed under the intermediate action is slower, due to the noisy nature of the intermediate action. From Fig.~\ref{fig:intermediate_speed} and \ref{fig:intermediate_loss}, we can see the trade-offs between the debiasing speed and the loss incurred, as making uniform exploration decisions could debias faster but it incurs a higher cumulative loss, as consistent with Theorems~\ref{thm:theta} and~\ref{thm:mdp}. 

\begin{figure}[ht]
\centering
	\subfigure
{\includegraphics[width=0.25\textwidth]{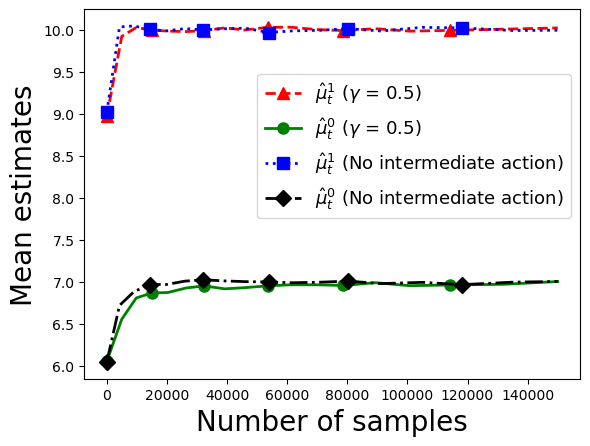}
	\label{fig:intermediate_performance}
	}
	\subfigure
	{	\includegraphics[width=0.25\textwidth]{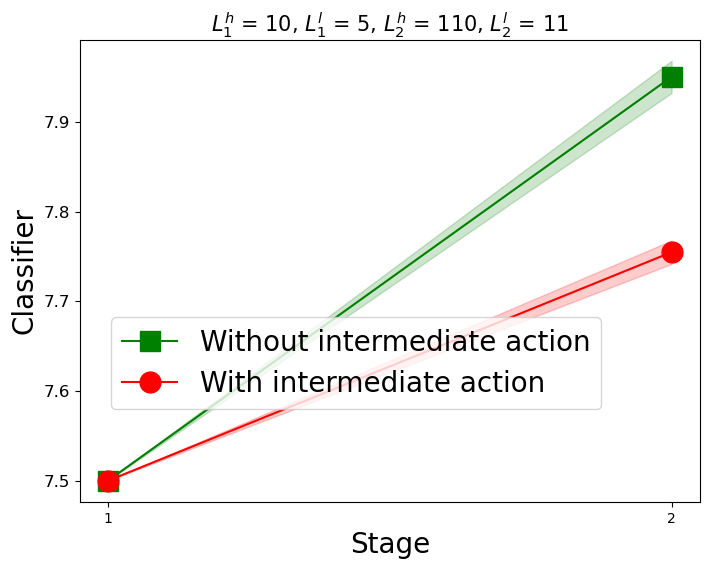}
		\label{fig:intermediate_speed}
	}
	\subfigure
	{
\includegraphics[width=0.25\textwidth]{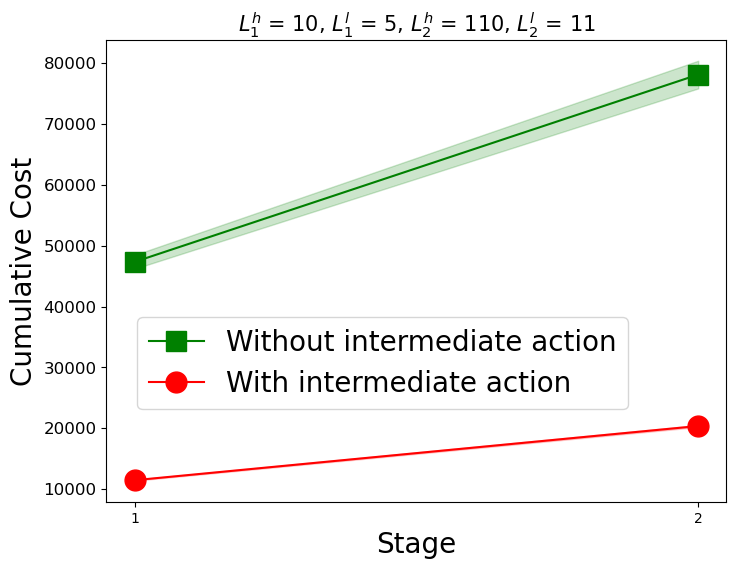}
		\label{fig:intermediate_loss}
	}
	\caption{Comparison with/out intermediate action} \label{fig:intermediate}
\end{figure}

\textbf{Intermediate actions in real-world datasets:} Fig.~\ref{fig:mdp-real} compares the impacts of opting for uniform vs. intermediate actions on the accumulated costs and debiasing speed in our two-stage MDP framework using real-world datasets: \emph{Adult} and \emph{FICO}. We let $\gamma = 0.5$, and we assume there are 500 samples sequentially arriving in each stage. We can see the trade-offs between the debiasing speed and the loss incurred, as making uniform exploration decisions could debias faster but it incurs a higher cumulative loss, as consistent with Theorems~\ref{thm:theta} and~\ref{thm:mdp}. 
\begin{figure}[ht]
 \centering
  \subfigure[Adult]{
\includegraphics[width=0.5\textwidth]{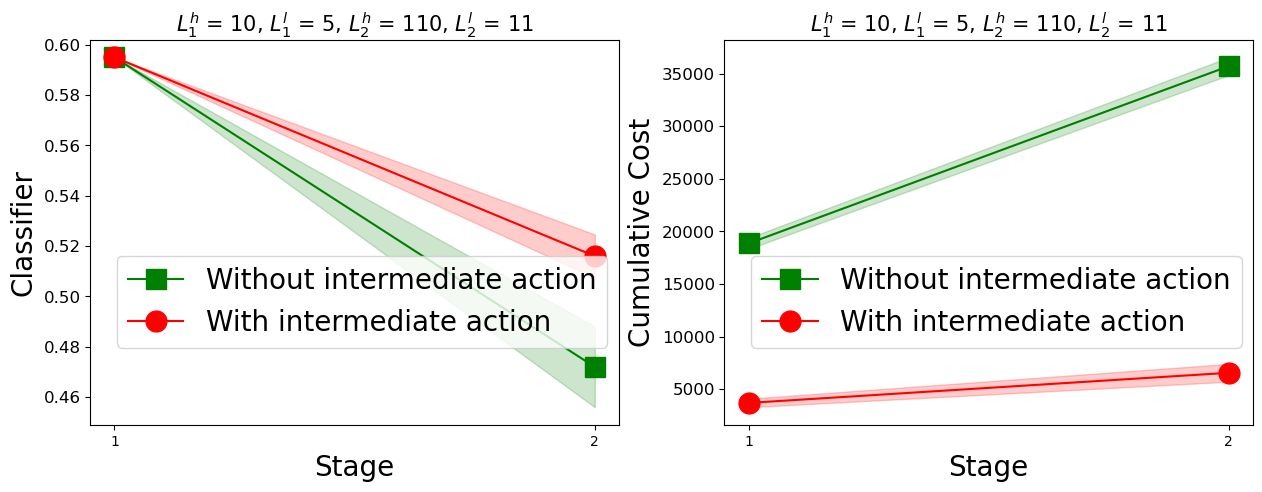}
 		\label{fig:adult_mdp}
 }%
 \subfigure[FICO]{
 \includegraphics[width=0.5\textwidth]{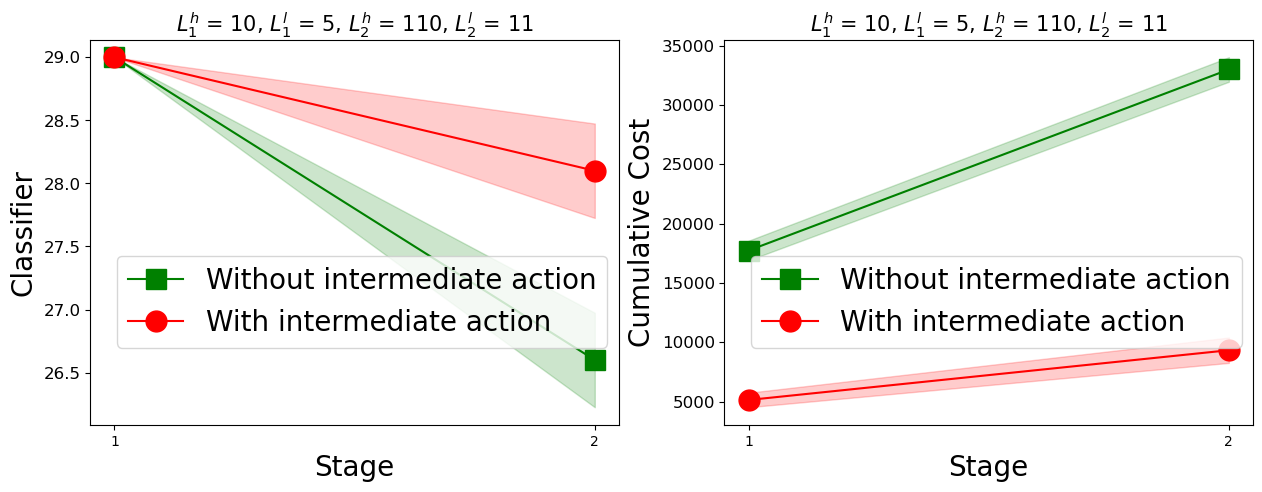}
 \label{fig:fico_mdp}
 }%
 \caption{Comparison with/out intermediate action via two-stage MDP using the real-world dataset.}
 \label{fig:mdp-real}
\end{figure}

\textbf{Interplay of debiasing and fairness constraints:} Fig.~\ref{fig:fairness-debiasing-real} compares the performance of our algorithm on real world data (e.g., \emph{Adult} and \emph{FICO}) when used in conjunction with three different fairness interventions: no fairness, equality of opportunity (EO), and the same decision rule (SD). The findings are consistent with Proposition 1.

\begin{figure}[ht]
 \centering
  \subfigure[Advantage label 0, Adult]{
\includegraphics[width=0.24\textwidth]{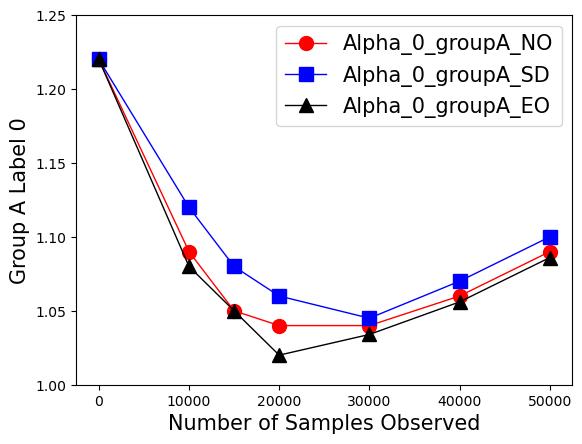}
 		\label{fig:adult_a01}
 }%
 \subfigure[Disadvantage label 0, Adult]{
 \includegraphics[width=0.24\textwidth]{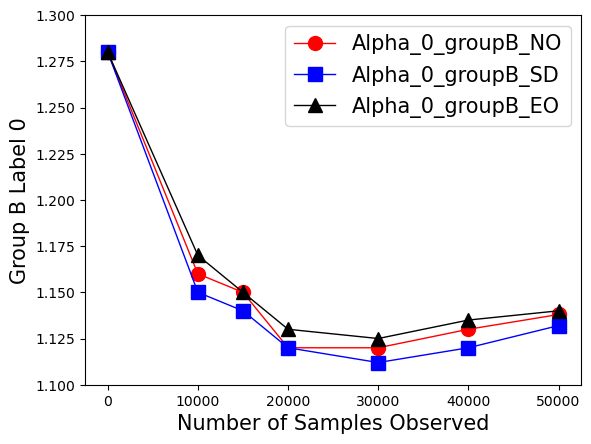}
 \label{fig:fico_a00}
 }%
 \subfigure[Advantage label 0, FICO]{
\includegraphics[width=0.24\textwidth]{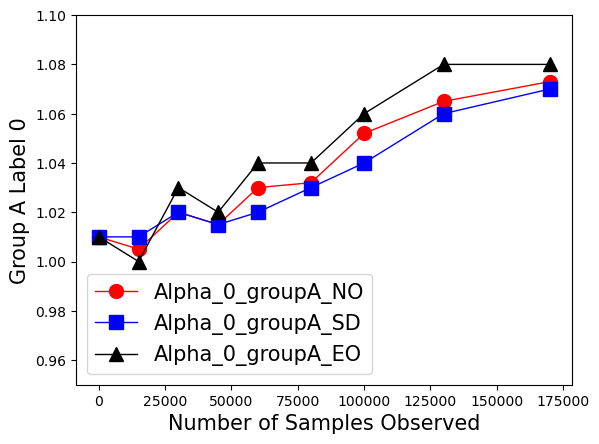}
 		\label{fig:fico_a01}
 }%
 \subfigure[Disadvantage label 0, FICO]{
 \includegraphics[width=0.24\textwidth]{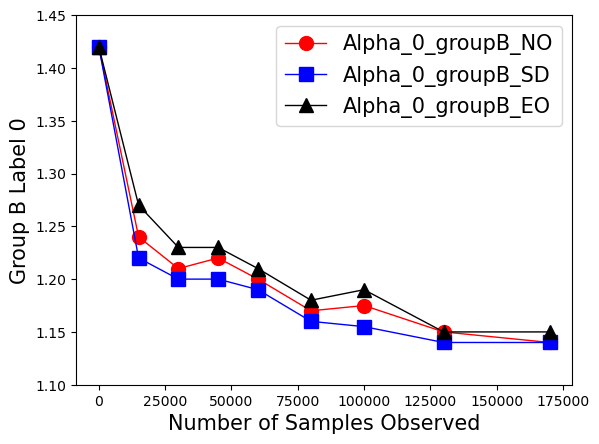}
 \label{fig:adult_a00}
 }%
 \caption{\texttt{Debiasing} with fairness constraints in real-world dataset.}
 \label{fig:fairness-debiasing-real}
\end{figure}

\textbf{Effects of depth of exploration:} Figure~\ref{fig:reference-points} compares the effects of modifying the depth of exploration through the choice of reference points on the performance of our \texttt{active debiasing} algorithm. In particular, we fix $\tau^1=50$ as the reference point on the qualified agents' estimates, and vary the reference points on unqualified agents' estimates in $\tau^0\in\{50,55,60\}$, with smaller reference points indicating deeper exploration (see Definition~\ref{def:UB-LB}). In all three settings, we reduce $\{\epsilon_t\}$ following a fixed reduction schedule. 

\begin{figure}[htbp!]
	\centering
 	\subfigure[False positives (unqualified agents admitted)]
 	{
		\includegraphics[width=0.41\textwidth]{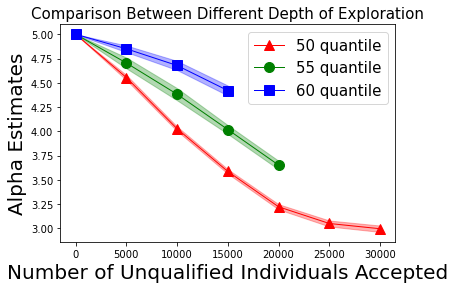}
		\label{fig:FP_reference-points}
	}%
	\hspace{0.05in}%
	\subfigure[False negatives (qualified agents rejected)]
	{
		\includegraphics[width=0.4\textwidth]{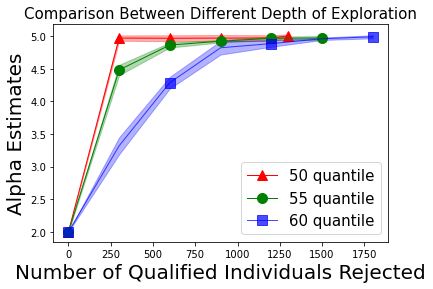}
		\label{fig:FN_reference-points}
	}%
	\caption{\texttt{Active debiasing} under different choices of depth of exploration, with $\tau^1=50$ and $\tau^0=\{50,55,60\}$. We reduce $\{\epsilon_t\}$ following a fixed reduction schedule.}
	\label{fig:reference-points}
\end{figure}

We first note that increasing the depth of exploration (here, e.g., setting $\tau^0=50$) leads to a faster speed of debiasing. This additional speed comes with a tradeoff: Fig.~\ref{fig:FP_reference-points} shows that algorithms with deeper exploration make more false positive errors, as they accept more unqualified individuals during exploration; by taking on this additional risk, they can debias the data faster. In addition, as observed in Fig.~\ref{fig:FN_reference-points}, the increased speed of debiasing means that the algorithm ultimately ends up making \emph{fewer} false negative decisions on the qualified individuals as a result of obtaining better estimates of their distributions. We conclude that a decision maker can use the choice of the reference point $\tau^0$ in our proposed algorithm to achieve their preferred tradeoff between the risk incurred due to incorrect admissions (higher FP) vs the benefit from the increased speed of debiasing and fewer missed opportunities (fewer FN). 
\section{Conclusion, Limitations, and Future Work}\label{sec:conclusion}

We proposed an \texttt{active debiasing} algorithm that introduces the idea of \emph{bounded} exploration as a way to limit exploration costs while recovering unbiased estimates of the underlying data distribution when future data suffers from censored feedback. We further explored the possibility of using bounded \emph{and noisy} exploration decisions (through the use of intermediate actions), and identified factors that can make the lower costs of noisy exploration worth the potential reduction in debiasing speed. We illustrated the performance of our proposed algorithms through numerical experiments on both synthetic and real-world datasets. Together, our findings highlight that our proposed algorithm's statistical/data debiasing effort can not only help improve the accuracy of the algorithm, but can also ultimately reduce social biases in the algorithm's decisions, shedding light on the importance of data debiasing in design responsible AI, and indicating a potential for alignment between accuracy (profit) and fairness goals in algorithmic decision making. 

\textbf{The single-unknown parameter assumption.} Our work focuses on learning of a single unknown parameter (Assumption~\ref{as:distributions}). Despite the commonality of this assumption in the multi-armed bandit learning literature, it also entails parametric knowledge of the underlying distribution with the other parameters such as variance or spread being known. We extend our algorithm to a Gaussian distribution with two unknown parameters in Online Appendix~\ref{app:two_parameters}. Extensions beyond this, especially those not requiring parametric assumptions on the underlying distributions, remain a main direction of future work. 

\textbf{On one-dimensional features and threshold classifiers.} Our analytical results have been focused on one-dimensional feature data and threshold classifiers. These assumptions may not be too restrictive in some cases: the optimality of threshold classifiers has been established in the literature by, e.g., \citep[Thm 3.2]{corbett2017algorithmic} and \citep{raab2021unintended}, as long as a multi-dimensional feature can be mapped to a properly defined scalar. Moreover, with recent advances deep learning, one can take the last layer outputs from a deep neural network and use it as the single dimensional representation. That said, any reduction of multi-dimensional features to a single-dimensional score may lead to some loss of information. In particular, our experiments have considered the use of our \texttt{active debiasing} algorithm on the \emph{Adult} dataset with multi-dimensional features by first performing a dimension reduction to a single-dimensional score; we find that this reduction can lead to a $\sim 5\%$ loss in performance. One potential solution to this is to adopt a mapping from high-dimensional features to scores that is revised repeatedly as the algorithm collects more data. Alternatively, one may envision a debiasing algorithm which targets its exploration towards collecting data on features that are believed to be highly biased; these remain as potential extensions of our algorithm. 

\textbf{Extensions of our analytical results.} We conjecture that Theorem~\ref{thm:debiasing} on the performance of \texttt{active debiasing} can be extended to distributions beyond unimodal distributions. Further, the analytical study of weighted regret of our algorithm, and comparison against the regret incurred by our two baselines, which we have observed numerically in Section~\ref{sec:simulations}, remain as main directions of future work. 

\textbf{Potential social impacts.} More broadly, while our debiasing algorithm imposes fairness constraints on its exploitation decisions (see problem~\eqref{eq:alg-obj}), it does not consider fairness constraints in its \emph{exploration} decisions. That means that our proposed algorithm could be disproportionate in the way it increases opportunities for qualified or unqualified agents in different groups during exploration. {Also, a limitation of our algorithm for groups with smaller representation is discussed in Section~\ref{sec:simulations}: in the \emph{Adult} dataset, as limited data is available on qualified, disadvantaged agents, the estimates on this population is not fully debiased. In other words, our algorithm is most effective at obtaining correct estimates on populations with sufficiently high representation. This may still be indirectly beneficial to the underrepresented populations, as by having better estimates on the represented population, the algorithm can better assess and impose fairness constraints.} That said, imposing fairness rules on exploration decisions, as well as identifying algorithms that can improve the speed of debiasing of estimates on underrepresented populations, can be explored to address these potential social impacts, and remain as interesting directions of future work.

\bibliographystyle{informs2014}
\bibliography{reference}

\clearpage

\setcounter{page}{1}
\setcounter{section}{0}
\section*{Online Supplements: Online Appendix}
\section{Proof of Theorem~\ref{thm:exploit-only_and_pure-exploration}}
We detail the proof of \texttt{exploitation-only} for $\hat{\parameter}^0_t$, and discuss two cases while assuming (wlog) that the unknown $\parameter^y$ being estimated is the distribution's mean. First, assume $\hat{\omega}^0_t$ is overestimated (i.e., $\hat{\omega}^0_t > \omega^0$). Note that we have $\theta_t \geq \hat{\omega}^0_t$. Then, as only agents with $x^\dagger\geq \theta_t$ are admitted, $\hat{\omega}^0_t$ may only be updated to stay the same or increase. Therefore, $\hat{\omega}^0_t$ will remain overestimated. 

Next consider the case that $\hat{\omega}^0_t$ is underestimated (i.e., $\hat{\omega}^0_t< \omega^0$). First, since each observation is independently drawn, we know that at time $t' = t, \ldots, t+T$, 
$
x_{t'} - \mathbb E[X|X \geq \theta_{t'}]
$
forms a martingale; this is because of the independence of $x_{t'}$ and $\theta_{t'}$ when conditioned on the historical information, as well as the fact that $\mathbb E[x_{t'}] = \mathbb E[X|X \geq \theta_{t'}]$. By definition of $\omega^0$, we also know that 
$
\sum_{t'=t}^T \mathbb E[X|X \geq \theta_{t'}]  > T \cdot \omega^0
$. 
Denote the gap by $\Delta:=\frac{\sum_{t'=t}^T \mathbb E[X|X \geq \theta_{t'}]}{T} -   \omega^0$. Therefore, using the Azuma-Hoeffding inequality we have 
\begin{align*}
    \mathbb P\Big(\sum_{t'=t}^T x_{t'} - \sum_{t'=t}^T \mathbb E[X|X \geq \theta_{t'}] \leq \delta \Big) \leq e^{\frac{-2\delta^2}{T-t+1}},
\end{align*}
for any $\delta < 0$. Letting $\delta = -\Delta \cdot (T-t+1)$, the above can be re-written as 
\begin{align*}
     \mathbb P&(\tfrac{1}{T-t+1} \sum_{t'=t}^T x_{t'} > \omega^0) >  1- e^{\left(-2\Delta^2(T-t+1)\right)} \underbrace{\rightarrow 1}_{T \rightarrow \infty}
\end{align*}
This proves that with high probability the mean of the new samples is higher than $\omega^0$. Therefore, at some time $T$ that is significantly higher than $t$, the new estimate $\hat{\omega}^0_T$ will be similar to $\frac{1}{T-t+1}\sum_{t'=t}^T x_{t'}$, which is higher than the true $\omega^0$. From our arguments for the overestimated case at the beginning of the proof, from this point on, $\hat{\omega}^0_t$ will stay overestimated. The proof for $\hat{\omega}^1_t$ is similar. 

For \texttt{pure exploration}, again assume (wlog) that the unknown parameter $\parameter^y$ being estimated is the distribution's mean. As we are collecting i.i.d. samples from across the distribution, $\hat{\parameter}^y_t$ can be set to the sample mean of the collected data-- more specifically, maintaining all data in the exploration region and down-sampling the data collected above the classifier with probability $\epsilon_t$, as also done in our proposed algorithm-- and the conclusion follows from the strong law of large numbers.

\section{Proofs of Theorem~\ref{thm:debiasing}}

We detail the proof for debiasing $\hat{f}_t^0$ (which happens using $x^\dagger \geq \text{LB}_t$ and $y^\dagger=0$); the proof for $\hat{f}_t^1$ is similar. 

\textbf{Part (a).} In time step $t+1$, with the arrival of a batch of $N_{t+1}$ samples in $[\text{LB}_t, \infty)$, the current estimate $\hat{\parameter}^0_t$ will be updated to $\hat{\parameter}^0_{t+1}$ based on the proportion of $\hat{\parameter}^0_t$ in the existing data. Denote the current left portion in $(\text{LB}_t, \hat{\parameter}^0_t)$ as $p_1 := \frac{\hat{F}^0(\hat{\parameter}^0_t) -\hat{F}^0(\text{LB}_t)}{\hat{F}^0(\theta_t)-\hat{F}^0(\text{LB}_t)}$. Based on Definition ~\ref{def:UB-LB}, we can also obtain the portion in $(\hat{\parameter}^0_t, \theta_t)$ denoted as $p_2 := \frac{\hat{F}^0(\theta_t) - \hat{F}^0(\hat{\parameter}^0_t)}{\hat{F}^0(\theta_t)-\hat{F}^0(\text{LB}_t)} = p_1$. We consider the following cases:

\underline{Case 1 (Perfectly estimated)}: $\hat{\parameter}_t^0=\parameter^{0}$. When the estimates are perfectly estimated, we will have both $\theta_t$ and $\text{LB}_t$ perfectly estimated too. Hence, we have $F^0(\theta_t) - F^0(\hat{\parameter}_t^0) = F^0(\hat{\parameter}_t^0) - F^0(\text{LB}_t)$ such that $p_1 = p_2$. Thus, $\mathbb{E}[\hat{\parameter}^0_{t+1}]=\parameter^0$. 
Hence, once the parameter is correctly estimated,  $\hat{f}_t^0$ is not expected to shift from $f^0$. 

\underline{Case 2 (Underestimated)}: $\hat{\parameter}_t^0<\parameter^{0}$. Under the unimodel distribution and single parameter assumption, since the arriving batch of data comes from the true distribution, $F^0(\text{LB}_t), F^0(\hat{\parameter}_t^0), F^0(\theta_t)$ will be smaller than $\hat{F}^0(\text{LB}_t), \hat{F}^0(\hat{\parameter}_t^0), \hat{F}^0(\theta_t)$, respectively. Moreover, since we have $\text{LB}_t \leq \hat{\parameter}_t^0 \leq \theta_t \leq \hat{\parameter}^1_t$, {then $F^0(\theta_t) - F^0(\hat{\parameter}_t^0) \geq F^0(\hat{\parameter}_t^0) - F^0(\text{LB}_t)$ such that $p_2 \geq p_1$. Hence, more samples are expected to be observed in range of $(\hat{\parameter}^0_t, \theta_t)$ so that the $\hat{\parameter}^0_t$ is expected to shift up.  Hence, we have $\mathbb{E}[\hat{\parameter}^0_{t+1}] \geq \hat{\parameter}^0_{t}$.} 

\underline{Case 3 (Overestimated)}: $\parameter^{0} <  \hat{\parameter}_t^0$.  Similar to \underline{Case 2 (Underestimated)}, we can obtain $\mathbb{E}[\hat{\parameter}^0_{t+1}] \leq \hat{\parameter}^0_{t}$.

\textbf{Part (b).} We first show that the converging sequence converges to the true estimates. 

By the construction of the bounds in Definition ~\ref{def:UB-LB}, the estimated parameter $\hat{\parameter}^0_{t}$ is the $\tau$-th percentile of $\hat{f}_t^0$, the median in the interval $[\text{LB}_t, \theta_t]$ and  some percentile in the interval $[\text{LB}_t, \infty)$; we therefore first find their distribution accordingly. Assume there are $N_t=m+n+1$ points in the interval $[\text{LB}_t, \infty)$ with $m$ and $n$ samples below and above $\hat{\parameter}^0_{t}$ respectively. More specifically, for these $n$ samples, there are $m$ samples between $[\hat{\parameter}^0_{t}, \theta_t]$ and $n-m$ samples above $\theta_t$. Based on the probability distribution of order statistics in $[\text{LB}_t, \theta_t]$, denote three possibilities $X$, $Y$, $Z$ denoting the number of samples below, on, and above the $\hat{\parameter}^0_{t}$, respectively, having probabilities $p= \tfrac{F^0(\hat{\parameter}^0_{t})-F^0(\text{LB}_t)}{F^0(\theta_t)-F^0(\text{LB}_t)}$, $q=\tfrac{f^0(\hat{\parameter}^0_{t})}{F^0(\theta_t)-F^0(\text{LB}_t)}$, and $r=\tfrac{F^0(\theta_t)-F^0(\hat{\parameter}^0_{t})}{F^0(\theta_t)-F^0(\text{LB}_t)}$.  Since the distributions are continuous, the probability of multiple samples being exactly on $\hat{\parameter}^0_{t}$ is zero. Therefore, the pdf of $\hat{\parameter}^0_{t}$ can be found based on the density function of the trinomial distribution: 
{\begin{align}
\mathbb P(\hat{\parameter}^0_{t}=\nu) \mathrm{d}\nu= \frac{(2m+1)!}{m!m!}(\tfrac{F^0(\nu)-F^0(\text{LB}_t)}{F^0(\theta_t)-F^0(\text{LB}_t)})^m(\tfrac{F^0(\theta_t)-F^0(\nu)}{F^0(\theta_t)-F^0(\text{LB}_t)})^m \tfrac{f^0(\nu)}{F^0(\theta_t)-F^0(\text{LB}_t)}\mathrm{d}\nu
\label{eq:median-pdf-trinomial}
\end{align}}
From the above, we can see that the density function of the $\hat{\parameter}^0_{t}$ is a beta distribution with $\alpha=m+1, \beta=m+1$, pushed forward by $H(\nu):=\tfrac{F^0(\nu)-F^0(\text{LB}_t)}{F^0(\theta_t)-F^0(\text{LB}_t)}$; this is the CDF of the truncated $F^0$ distribution in $[\text{LB}_t, \theta_t]$. In other words, using $G$ to denote the Beta distribution's CDF, $\hat{\parameter}^0_{t}$ has CDF $G(H(\nu))$, and by the chain rule, pdf $g(H(\nu))h(\nu)$.  

It is known ~\cite{maritz1978note} that for samples located in the range of $[\text{LB}_t, \theta_t]$, the sampling distribution of the median becomes asymptotically normal with mean $(\parameter^0)'$ and variance $\frac{1}{4(2m+3)H((\parameter^0)')}$, where $(\parameter^0)'$ is the median, the truncated $F^0$ distribution in $[\text{LB}_t, \theta_t]$. If the sequence of $\{\hat{\parameter}^0_t\}$ produced by our \texttt{active debiasing} algorithm converges, by Definition ~\ref{def:UB-LB}, the thresholds $\text{LB}_t$ and $\theta_t$ will converge as well; As $t\rightarrow \infty$, $\epsilon_t \rightarrow 0$, $2m+1\rightarrow\infty$ in this interval, the variance becomes zero, and  $\hat{\parameter}^0_{t+1}\rightarrow (\parameter^0)'$. By Definition ~\ref{def:UB-LB}, it must be that the median $(\parameter^0)'$ of $H$ is equal to $\parameter^0$. Therefore, $\hat{\parameter}^0_{t+1}\rightarrow \parameter^0$.

Lastly, we show that the sequence of estimates $\{\hat{\parameter}^0_t\}$ is a converging sequence. Consider the sequence of estimates $\{\hat{\parameter}^0_t\}$, and separate into the two disjoint subsequences $\{\hat{y}^0_t\}$ denoting the parameters that are underestimated with respect to the true $\parameter^0$, and $\{\hat{z}^0_t\}$ denoting those that are overestimated.

We now show the sequence of underestimation errors, $\{\Delta^y_t\} : = \{\parameter^0- \hat{y}^0_t\}$ and the sequence of overestimation errors, $\{\Delta^z_t\} : = \{\hat{z}^0_t-\parameter^0\}$, are supermartingales. We detail this for $\{\Delta^y_t\}$. Consider two cases: 
\begin{itemize}
    \item First, assume the update $\hat{y}^0_{t+1}$ is the next immediate update after $\hat{y}^0_{t}$ in the original sequence $\{\hat{\parameter}^0_t\}$; that is, an underestimated $\hat{y}^0_{t}$ has been updated to a parameter that continues to be an underestimate. In this case, by Part (a), $\mathbb{E}[\hat{y}^0_{t+1}|\hat{y}^0_{t}]\geq \hat{y}_t^0$, and therefore, $\mathbb{E}[\Delta^y_{t+1}|\Delta^y_{t}]\leq \Delta^y_t$. 
    \item Alternatively assume $\hat{y}^0_{t+1}$ is not obtained immediately from $\hat{y}^0_{t}$; that is, $\hat{y}^0_{t+1}$ has been obtained as a result of an update from an overestimated parameter. We note that now, $\hat{y}^0_{t+1}\geq \hat{y}_t^0$. This is because either no new estimates have been obtained between $\hat{y}_t^0$ and the true parameter $\parameter^0$ since the last time the parameter was underestimated, in which case, it must be that $\hat{y}^0_{t+1}=\hat{y}_t^0$. Otherwise, a new estimate in $[\hat{y}_t^0, \parameter^0]$ has been obtained, in which case, again, $\mathbb{E}[\hat{y}^0_{t+1}|\hat{y}^0_{t}]\geq \hat{y}_t^0$. In either case, $\mathbb{E}[\Delta^y_{t+1}|\Delta^y_{t}]\leq \Delta^y_t$. 
\end{itemize}

Therefore, by the Doobs Convergence theorem, the supermatingales $\{\Delta^y_t\}$ and $\{\Delta^z_t\}$ converge to random variables $\Delta^y$ and $\Delta^z$. By the same argument as the beginning of the proof of this part, these are asymptotically normal with mean zero and with variances decreasing in the number of observed samples in their respective intervals. Therefore, $\Delta^y\rightarrow 0$ and $\Delta^z\rightarrow 0$ as $N\rightarrow \infty$, and therefore $\{\hat{\parameter}^0_t\}$ converges to $\parameter$.

\section{Proof of Theorem~\ref{thm:regret}}\label{app:regret_proof}
Our algorithm's error is made up of errors from four different sources, which we characterize step by step. Firstly, we have an \emph{approximately} optimal classifer that is returned by the exponentiated gradient algorithm with suboptimiality level $v$ (step 1). Secondly, we use samples to estimate the distributions and thus have empirical biases (step 2). Thirdly, since we start from biased distributions, there are errors due to the domain mismatches (step 3). Lastly, in order to debias, we explore by admitting samples that would otherwise be rejected, introducing additional errors (step 4). 

Before proceeding, we outline our notation for different forms of data bias. First, there is a true underlying distribution for the population of agents to which the classifier is to be applied; we denote this by $\bar{D}$. Our focus in this work is on setting where there are different forms of statistical biases in the training data (e.g. distribution shifts or adaptive sampling biases); denote this statistically biased training data by $\tilde{D}$. Finally, even without distribution shifts or adaptive sampling biases, the classifier has access to a limited, empirically biased subset of this data; we denote the initial statistically and empirically biased data distribution by $\hat{D}$. 

Accordingly, let $\hat{h}^*_{\theta_{g,0}}, \tilde{h}^*_{\theta_{g,0}}, \bar{h}^*_{\theta_{g}}$ be the optimal (fair and error minimizing) classifiers that would be obtained from an initial statistically and empirically biased dataset, only statistically biased dataset, and an unbiased dataset,  respectively. 

\noindent \textbf{Step 1: Approximate solution errors.} We can treat the problem of finding the initial fair classifier from the statistically and empirically biased training data as a saddle point problem. First, let $\tilde{err}(h_{\theta_{g,t=0}})=\displaystyle \mathop{\mathbb{E}}_{(x_i,y_i,g_i)\sim \tilde{D}}\Big[\ell(h_{\theta_{g,t=0}}(x_i,g_i),y_i)\Big]$; this is the true error incurred by a classifier $h_{\theta_{g,t=0}}$ when training data comes from $\tilde{D}$, and is the objective function of the minimization problem. Additionally, we assume throughout that a fairness constraint $|\mathcal{C}(\theta_{a,t}, \theta_{b,t})| \leq \gamma$ has been imposed.  

However, since we do not have the true $\tilde{D}$, and only have access to a limited, empirically biased subset of it $\hat{D}$, we will use the empirical estimates $\hat{err}(h_{\theta_{g,0}}), \hat{\mathcal{C}}(\theta_{a,0}, \theta_{b,0})$ and $\hat{\gamma}$ in the constrained optimization problem of finding the fair, loss-minimizing classifier. To capture the fairness constraint, we will introduce Lagrangian multipliers $\lambda_j \geq 0$. This allows us to define the Lagrangian of the optimization problem:
\begin{align*}
\mathcal{L}(h_{\theta_{g,0}},\lambda_j)= \hat{err}(h_{\theta_{g,0}}) + \lambda_1 (\hat{\mathcal{C}}(\theta_{a,0}, \theta_{b,0})-\hat{\gamma}) + \lambda_2 (-\hat{\mathcal{C}}(\theta_{a,0}, \theta_{b,0})-\hat{\gamma})
\end{align*}
{Following the rewriting procedures in \citet{agarwal2018reductions}} and using the exponentiated gradient algorithm, we can obtain a $v$-approximated solution $(\hat{h}_{\theta_{g,0}},\hat{\lambda}_j)$; this is an approximately loss-minimizing fair classifier obtained based on an initial, empirically and statistically-biased training data, and the corresponding Lagrange multipliers of the fairness constraint.  

\noindent \textbf{Step 2: Empirical error bound on the initial biased distribution.} To bound the statistical error, we use Rademacher complexity of the classifier family $\mathcal{H}$ denoted as $\mathcal{R}_n(\mathcal{H})$, where $n$ is the number of training samples. Let $n_{g,t}$ be the number of training samples arriving in round $t$ from agents in group $g$. Initially, we have $n_{g,0}= b^0_{g,0}+b^1_{g,0}$. We also assume that $\mathcal{R}_n(\mathcal{H}) \leq Cn^{-\alpha}$ for some $C\geq 0$ and $\alpha \leq 1/2$. Hence, based on the Theorem 4 in \citet{agarwal2018reductions}, we can find that with probability at least $1 - 4\delta$ with $\delta >0$:
\begin{equation}
    \tilde{err}(\hat{h}_{\theta_{g,0}}) \leq \tilde{err}(\tilde{h}^*_{\theta_{g,0}}) +2v + 4\mathcal{R}_{n_{g,0}}(\mathcal{H}) + \frac{4}{\sqrt{n_{g,0}}} + \sqrt{\frac{2\ln(2/\delta)}{n_{g,0}}} 
    \label{eq:empirical_error}
\end{equation}
In words, this provides a bound on the true error that will be incurred on statistically biased data when using the classifier obtained in step 1 (from statistically and empirically biased data). 

\noindent \textbf{Step 3: Bound of error on different distributions (domains).} Next, we note that there is a mismatch between our current biased training data and the true underlying data distribution. We use results from domain adaptation to bound these errors. 

To bound the error on different distributions, $L^1$ divergence would be a nature measure. However, it overestimates the bounds since it involves a supremum over all measurable sets. As discussed by \citet{ben2010theory}, using classifier-induced divergence ($\mathcal{H}\Delta\mathcal{H}$-divergence) allows us
to directly estimate the error of a source-trained classifier on the target domain by representing errors relative to other hypotheses. 

\begin{definition}[$\mathcal{H}\Delta\mathcal{H}$-divergence]
For a hypothesis space $\mathcal{H}$, the symmetric difference hypothesis space $\mathcal{H}\Delta\mathcal{H}$ is
the set of hypotheses:
\[g \in \mathcal{H}\Delta\mathcal{H} \Leftrightarrow g(x) = h(x) \oplus h'(x) \hspace{0.1in} \text{for some } h, h' \in \mathcal{H}\]
where $\oplus$ is XOR function. In other words, every hypothesis $g$ is the set of disagreement between two hypotheses in $\mathcal{H}$. The $\mathcal{H}\Delta\mathcal{H}$-distance is also given by
\[d_{\mathcal{H}\Delta\mathcal{H}}(D,D') = 2 \sup_{h,h' \in \mathcal{H}} \Big|Pr_{x\sim D}[h(x) \neq h'(x)] - Pr_{x\sim D'}[h(x) \neq h'(x)] \Big|\]
\end{definition}

Let $\overline{err}(h)$ be the error made by a classifier $h$ on unbiased data from the true underlying distribution. We bound this error below. 
\begin{lemma}[Follows from Theorem 2 of \citet{ben2010theory}]
Let $\mathcal{H}$ be a hypothesis space. If unlabeled samples are from $\tilde{D}_{g,0}$ and $D_g$ respectively, then for any $\delta \in (0,1)$, with
probability at least $1 - \delta$:
\begin{equation}
    \overline{err}(\hat{h}_{\theta_{g,0}}) \leq \tilde{err}(\hat{h}_{\theta_{g,0}}) + \tfrac{1}{2}d_{\mathcal{H}\Delta\mathcal{H}}(\tilde{D}_{g,0}, D_g)+c(\tilde{D}_{g,0}, D_g) 
    \label{eq:source_target}
\end{equation}
where $\tilde{D}_{g,0}$ and $D_g$ are the joint distribution of labels, and $c(\tilde{D}_{g,0}, D_g) = \min_h \overline{err}(h) + \tilde{err}(h)$.
\end{lemma}

Then, combining equation ~\ref{eq:empirical_error} and ~\ref{eq:source_target}, we can obtain the following expression:
\begin{align*}
    &\overline{err}(\hat{h}_{\theta_{g,0}}) \leq  \tilde{err}(\hat{h}_{\theta_{g,0}}) + \tfrac{1}{2}d_{\mathcal{H}\Delta\mathcal{H}}(\tilde{D}_{g,0}, D_g)+c(\tilde{D}_{g,0}, D_g) \\
    &\qquad \leq  \tilde{err}(\tilde{h}^*_{\theta_{g,0}}) +2v + 4\mathcal{R}_{n_{g,0}}(\mathcal{H}) + \tfrac{4}{\sqrt{n_{g,0}}} + \sqrt{\tfrac{2\ln(2/\delta)}{n_{g,0}}} + \tfrac{1}{2}d_{\mathcal{H}\Delta\mathcal{H}}(\tilde{D}_{g,0}, D_g)+c(\tilde{D}_{g,0}, D_g) \\
    &\qquad \leq  \tilde{err}(\bar{h}^*_{\theta_{g,0}}) +2v + 4\mathcal{R}_{n_{g,0}}(\mathcal{H}) + \tfrac{4}{\sqrt{n_{g,0}}} + \sqrt{\tfrac{2\ln(2/\delta)}{n_{g,0}}} + \tfrac{1}{2}d_{\mathcal{H}\Delta\mathcal{H}}(\tilde{D}_{g,0}, D_g)+c(\tilde{D}_{g,0}, D_g) \\
    &\qquad \leq  \overline{err}(\bar{h}^*_{\theta_{g,0}}) +2v + 4\mathcal{R}_{n_{g,0}}(\mathcal{H}) + \tfrac{4}{\sqrt{n_{g,0}}} + \sqrt{\tfrac{2\ln(2/\delta)}{n_{g,0}}} + d_{\mathcal{H}\Delta\mathcal{H}}(\tilde{D}_{g,0}, D_g)+2c(\tilde{D}_{g,0}, D_g) 
\end{align*}
In words, this provides a bound on the true error that will be incurred on the unbiased data from the underlying population when using the classifier obtained in step 1 (from statistically and empirically biased data). 

\noindent \textbf{Step 4: Exploration errors.} Lastly, in order to reduce the mismatches between the biased training data and the true underlying distribution, our algorithm incurs some exploration errors. Let $n'_{0,g,t}$ and $n'_{1,g,t}$ denote the number of samples from unqualified and qualified group that fall below the threshold $\theta_{g,t}$ in round $t$, respectively. Since in Steps 2 and 3 we already considered the classification errors due to empirical estimation and different distributions, we only consider the additional exploration error introduced with the goal of removing biases. Because of exploration, some qualified samples that were rejected previously will now be accepted, which will allow the algorithm to make fewer errors. Similarly, some unqualified samples that would previously be rejected are now accepted, which will lead to an increase in errors.

Denote $\epsilon_t$ as the exploration probability at round $t$. The exploration error consists of the errors made on the unqualified group, minus correct decisions made on the qualified group. In bounded exploration approach, we introduce a $\text{LB}_t$ to limit the depth of exploration. Therefore, the number of samples that fall into the exploration range will be proportional to $n'_{0,g,t}$ and $n'_{1,g,t}$ based on the location of $\text{LB}_t$. Mathematically, denote $N_{g,t}$ as the net exploration error for group $g$ at round $t$; this is given by:
\[N_{g,t} := \Bigg(\frac{\hat{F}^0_{g,t}(\theta_t)-\hat{F}^0_{g,t}(\text{LB}_t)}{\hat{F}^0_{g,t}(\theta_t)}\epsilon_tn'_{0,g,t} - \frac{\hat{F}^1_{g,t}(\theta_t)-\hat{F}^1_{g,t}(\text{LB}_t)}{\hat{F}^1_{g,t}(\theta_t)}\epsilon_tn'_{1,g,t}\Bigg)\]

\noindent \textbf{Step 5: Errors made over $m$ updates.} We now state the error incurred by our algorithm over $m$ rounds of updates. For a group $g$, combining the four identified sources of error over $m$ updates, we have
{\small
\begin{align*}
    \sum_{t=1}^{m}\overline{err}(\hat{h}_{\theta_{g,t}}) \leq \sum_{t=1}^{m} \Big[\overline{err}(\bar{h}^*_{\theta_{g,t}}) + 2v + 4\mathcal{R}_{n_{g,t}}(\mathcal{H}) + \tfrac{4}{\sqrt{n_{g,t}}} + \sqrt{\tfrac{2\ln(2/\delta)}{n_{g,t}}} + N_{g,t} + d_{\mathcal{H}\Delta\mathcal{H}}(\tilde{D}_{g,t}, D_g)+2c(\tilde{D}_{g,t}, D_g) \Big]
\end{align*}
}
Therefore, the error bound for our algorithm over $m$ updates and across two groups $g\in\{a,b\}$ is given by 
{
\begin{align*}
&\text{Err.} = 
    \sum_{t=1}^{m}\Big[\overline{err}(\hat{h}_{\theta_{a,t}}) + \overline{err}(\hat{h}_{\theta_{b,t}}) - \overline{err}(h^*_{\theta_{a,t}}) - 
    \overline{err}(h^*_{\theta_{b,t}}) \Big] \\
    &\qquad \hspace{-0.3in} \leq \sum_{g,t} \Big[ \underbrace{2v}_{\text{$v$-approx.}} + \underbrace{4\mathcal{R}_{n_{g,t}}(\mathcal{H}) + \tfrac{4}{\sqrt{n_{g,t}}} + \sqrt{\tfrac{2\ln(2/\delta)}{n_{g,t}}}}_{\text{empirical estimation}} + \underbrace{N_{g,t}}_{\text{explor.}} + \underbrace{d_{\mathcal{H}\Delta\mathcal{H}}(\tilde{D}_{g,t}, D_g)+2c(\tilde{D}_{g,t}, D_g)}_{\text{source-target distribution}} \Big] 
\end{align*}}

From the expression above, we can see that the error incurred by our algorithm consists of four types of error: errors due to approximation of the optimal (fair) classifier at each round, empirical estimation errors, exploration errors, and errors due to our biased training data (viewed as source-target distribution mismatches); the latter two are specific to our \texttt{active debiasing} algorithm. In particular, as we collect more samples, $n_{g,t}$ will increase. Hence, the empirical estimation errors decrease over time. Moreover, as the mismatch between $\tilde{D}_{g,t}$ and $D_g$ decreases using our algorithm (by Theorem~\ref{thm:debiasing}), the error due to target domain and source domain mismatches also decrease. In the meantime, our exploration probability $\epsilon_t$ also becomes smaller over time, decreasing $N_{g,t}$. Together, these mean that the terms in the summation above decrease as $t$ increases. 

\section{Proof of Proposition~\ref{prop:fairness-debiasing}} \label{app:proof_fairness_debiasing}
We compare the speed of debiasing through $\mathbb{E}[|\hat{\parameter}^y_t-\parameter^y|]$. Given a fixed $t$, we say the algorithm for which this error is larger has a lower speed of debiasing. In other words, the slower algorithm needs to wait for \emph{more} arriving samples before it can reach the same parameter estimation error as a faster algorithm. 

We prove the proposition for the case where the introduction of fairness constraints leads to over-selection of group $g$, i.e., $\theta^{F}_{g,t}<\theta^{U}_{g,t}$. The proofs for the under-selected case are similar. We note that the presence of two different groups only affects the choice of the classifier given the fairness constraints, following which the proof becomes independent of the group label; we therefore drop $g$ in the remainder of the proof.

We detail the proof for the debiasing of $\hat{f}^0_{t}$, which depends on the choice of $\text{LB}_t$ in Definition~\ref{def:UB-LB}, i.e., 
$\hat{F}^{0}_t(\text{LB}_t) = 2\hat{F}^{0}_t(\hat{\parameter}^0_t)-\hat{F}^{0}_t(\theta_t)~.$
Since $\theta^{F}_{t}<\theta^{U}_{t}$, this means that $\hat{F}^{0}_t(\theta^{F}_{t})<\hat{F}^{0}_t(\theta^{U}_{t})$, and consequently that $\hat{F}^{0}_t(\text{LB}^F_t)>\hat{F}^{0}_t(\text{LB}^U_t)$, and thus, that $\text{LB}^F_t>\text{LB}^U_t$.

Now, consider the interval $[\text{LB}_t, \max^0]$, with $\max^0$ denoting the maximum of ${f}^0$. Only arrivals of $(x^\dagger, y^\dagger)$, with $y^\dagger=0$, who are admitted in this interval, will result in an update to the estimated median. Since $\text{LB}^F_t>\text{LB}^U_t$, this interval is narrower under the fairness constrained classifier, meaning that it takes more time to meet the batch size requirement under compared $\text{LB}^U_t$ compared to $\text{LB}^F_t$. As detailed in the proof of Theorem~\ref{thm:debiasing} each of these updates will move the estimate in the correct direction, and these estimates converge to the true value in the long-run as more samples become available. Hence, debiasing of $\hat{f}_t^0$ is slower after the introduction of fairness constraints. 

Similar arguments hold for updating $\hat{f}^1_{t}$, which takes samples in $[\text{LB}_t, \max^1]$. When $\text{LB}_t$ increases, it also takes more time for label 1 distribution update. Hence, after the introduction of the constraint, the fairness unconstrained classifier observes a wider range of sample points, including all those observed by the constrained classifier. Therefore, fairness constraints decrease the speed of debiasing on $\hat{f}_t^1$ as well.

\section{Proof of Theorems~\ref{thm:theta} and \ref{thm:mdp}} \label{app:proof_mdp}
Our analysis consists of four steps. Step 1: For given $\hat{f}^1$ and $\hat{f}^0$, we derive for the loss-minimizing classifier $\hat{\theta}$ according to Eq.~\ref{eq:alg-obj}. Step 2: With newly collected samples in hand, we derive the closed form for the updated distribution estimates. Step 3: We combine Steps 1 and 2 to illustrate the impact of intermediate action on the debiasing speed. Step 4: Integrating different penalty weights for various decisions, we demonstrate the impact of intermediate action on the cumulative cost. Without loss of generality, we assume the initial biased estimates are underestimated.

Step 1: According to the Eq.~\ref{eq:alg-obj} and distribution assumption, by the first-order optimality condition, we could obtain $\alpha^1 \hat{f}^1(\hat{\theta}) - \alpha^0 \hat{f}^0(\hat{\theta}) = 0$, and $\hat{\theta} = \frac{\hat{\mu}^1 + \hat{\mu}^0}{2} - \frac{\sigma^2\log(\alpha^1/\alpha^0)}{\hat{\mu}^1 - \hat{\mu}^0}$

Step 2: Since the updates for the label 1 and label 0 distribution estimates follow the same procedure, our analysis primarily focuses on the update process for the label 1 distribution. Suppose we initially have $n$ biased samples for the label 1 distribution, and $N_1$ samples arrive at $t=1$. Among these, $m^U$ and $m^{I}$ samples are collected by choosing $A = \{U, I\}$, respectively. Due to the unreliability of explored labeled 1 samples, we have $m^{I} \leq m^U$ in essence. With newly collected $m$ samples, the updated $\hat{\mu^y}$ could be calculated as follows:
    $\hat{\mu}^y_{t+1} = \frac{x_1 + \dots + x_n + x'_1 + \dots + x'_m}{n+m} = \frac{n}{n+m} \hat{\mu}^y_t + \frac{x'_1 + \dots + x'_m}{n+m} $
{\remark{ The derivation is based on a single calculation involving a batch of $m$ samples. The derivation remains unchanged if we conduct $m$ calculations, each involving an individual sample.}}

Based on our assumptions and Theorem~\ref{thm:debiasing}, we can view the extra samples in $m^U$ compared to $m^{I}$ as further updates to the distribution estimates. Consequently, we have $\mathbb{E}[\hat{\mu}^{y,U}_2] \geq \mathbb{E}[\hat{\mu}^{y, I}_2]$ for any $y \in \{0,1\}$.

Step 3: For simplicity in this analysis, we set a UB similar to LB, and consider $\hat{w}^0$ is the median of $\hat{f}^0$. We then follow Algorithm~\ref{def:db-alg}'s procedure with the same type of exploitation and exploration decisions. 
\begin{definition}\label{def:UB}
    At time $t$, the firm selects a upper bound $\text{UB}_t$ such that
    \[UB_t = (\hat{F}^{1}_t)^{-1}(2\hat{F}^{1}_t(\hat{\mu}^1_t)-\hat{F}^{1}_t(\text{LB}_t)), \hspace{0.2in}
    UB'_t = (\hat{F}^{1}_t)^{-1}(2\hat{F}^{1}_t(\hat{\mu}^1_t)-\hat{F}^{1}_t(\hat{\theta}_t)) \]
    where $\text{LB}_t$ is obtained from Definition \ref{def:UB-LB}, $\hat{\theta}_t$ is the current loss-minimizing classifier, $\hat{F}_t^1$, $(\hat{F}^{1}_t)^{-1}$ are the cdf and inverse cdf of the estimated distribution $\hat{f}_{t}^1$, respectively. 
\end{definition}
Based on Steps 1 and 2, we could write the expression for $\hat{\theta}^U_2$ and $\hat{\theta}^{I}_2$, respectively.
    \[\hat{\theta}^U_2 = \frac{\hat{\mu}^{1,U}_2 + \hat{\mu}^{0,U}_2}{2} - \frac{\sigma^2\log(\alpha^1/\alpha^0)}{\hat{\mu}^{1,U}_2 - \hat{\mu}^{0,U}_2}, \hspace{0.2in} \hat{\theta}^{I}_2 = \frac{\hat{\mu}^{1,I}_2 + \hat{\mu}^{0,I}_2}{2} - \frac{\sigma^2\log(\alpha^1/\alpha^0)}{\hat{\mu}^{1,I}_2 - \hat{\mu}^{0,I}_2}\]
We analyze the differences between $\hat{\theta}^U_2$ and $\hat{\theta}^{I}_2$ term by term. Regarding the first term, it is clearly that $\frac{\hat{\mu}^{1,U}_2 + \hat{\mu}^{0,U}_2}{2} \geq \frac{\hat{\mu}^{1,I}_2 + \hat{\mu}^{0,I}_2}{2}$ because of $\hat{\mu}^{y,U}_2 \geq \hat{\mu}^{y, I}_2$ for any $y \in \{0,1\}$. Concerning the second term, it's important to note that the uncertainty introduced by the intermediate action causes a reduction in the range of labeled 1 samples used for updates, from $(LB_2, UB_2)$ to $(\hat{\theta}_2, UB'_2)$, where $LB_2 \leq \hat{\theta}_2$ and $UB_2 \geq UB'_2$. Furthermore, although the range of labeled 0 samples used for updates remains the same, only samples successfully fulfilling the requirements of the intermediate action could be used. 
    
In the reduced sample set, the number of labeled 0 samples would decrease by the number of samples falling within the range $(LB_2, \hat{\theta}_2)$ with an exploration probability $\epsilon$ and fulfilling probability $\gamma$. Conversely, the number of labeled 1 samples would decrease by the number of samples in the range $(LB_2, \hat{\theta}_2)$ and $(UB'_2, UB_2)$, which reduces more "diverse" samples compared to labeled 0 samples. Therefore, as also shown in our experiment Fig~\ref{fig:base_over}, the change in $\hat{\mu}^1$ is greater than that in $\hat{\mu}^0$ such that 
$\mathbb{E}[\mu^{1, U}_2 - \mu^{1, I}_2] \geq \mathbb{E}[\mu^{0, U}_2 - \mu^{0, I}_2]$. Hence, we can get $\mathbb{E}[\mu^{1, U}_2 - \mu^{0, U}_2] \geq \mathbb{E}[\mu^{1, I}_2 - \mu^{0, I}_2]$. The analysis becomes straightforward when $\alpha^1=\alpha^0$ because the second term is 0. Therefore, we have $\mathbb{E}[|\theta^* - \hat{\theta}^U_2|] \leq \mathbb{E}[|\theta^* - \hat{\theta}^{I}_2|]$.

Step 4: When $t=1$, before taking any actions, the loss-minimizing classifier $\hat{\theta}_1$ is obtained through an initial biased training dataset. Therefore, according to the derived expression, we can find the differences between misclassification costs would be 0. We can also write the expression for the difference between the expected exploration cost such that  
   {\footnotesize \begin{align*}
        & \mathbb{E}[\mathcal{L}^{\text{exp-cost}}_1(U, \hat{f}_1^0, \hat{f}_1^1)] - \mathbb{E}[\mathcal{L}^{\text{exp-cost}}_1(I, \hat{f}_1^0, \hat{f}_1^1)] \\
        &\qquad \hspace{-0.2in}= N_1 \Big \{ -L^h_1 \alpha^1 \int_{LB_1}^{\hat{\theta}_1} f^1(x)dx + L^h_2  \alpha^0 \int_{LB_1}^{\hat{\theta}_1} f^0(x)dx\Big \} \\
        &\qquad \hspace{1.5in} - N_1 \Big \{ (-L^h_1+L^l_1)\alpha^1 \int_{LB_1}^{\hat{\theta}_1} f^1(x)dx + L^l_2(1-\gamma)  \alpha^0 \int_{LB_1}^{\hat{\theta}_1} f^0(x)dx\Big \}\\
        &\qquad \hspace{-0.2in} = N_1 \Big\{ - L^l_1\alpha^1 \int_{LB_1}^{\hat{\theta}_1} f^1(x)dx +  L^h_2  \alpha^0 \int_{LB_1}^{\hat{\theta}_1} f^0(x)dx - L^l_2(1-\gamma) \alpha^0 \int_{LB_1}^{\hat{\theta}_1} f^0(x)dx\Big\}\\
        &\qquad \hspace{-0.2in} \geq N_1 \Big\{ - L^l_1\alpha^1 \int_{LB_1}^{\hat{\theta}_1} f^0(x)dx +  L^h_2  \alpha^0 \int_{LB_1}^{\hat{\theta}_1} f^0(x)dx - L^l_2(1-\gamma) \alpha^0 \int_{LB_1}^{\hat{\theta}_1} f^0(x)dx\Big\}
    \end{align*}}
The inequality is derived from our assumption regarding the mode of $f^1$ being greater than that of $f^0$. 
    
In the two-stage MDP framework, the remaining terms are the misclassification cost at $t=2$. At $t=2$, according to our expressions, we could write it as follows:
    \begin{align*}
        &\mathbb{E}[\mathcal{L}^{\text{miss-cost}}_t(U, \hat{f}_t^0, \hat{f}_t^1))] - \mathbb{E}[\mathcal{L}^{\text{miss-cost}}_t(I, \hat{f}_t^0, \hat{f}_t^1))] \\
        &\qquad = N_2 \Big \{L^h_1 \alpha^1 \int_{-\infty}^{\hat{\theta}^U_2} f^1(x)dx + L^h_2 \alpha^0 \int_{\hat{\theta}^U_2}^{\infty} f^0(x)dx\Big\} \\
        &\qquad \hspace{2in}- N_2 \Big \{L^h_1 \alpha^1 \int_{-\infty}^{\hat{\theta}^{I}_2} f^1(x)dx + L^h_2 \alpha^0 \int_{\hat{\theta}^{I}_2}^{\infty} f^0(x)dx\Big\} \\
        &\qquad = N_2\Big \{ L^h_1 \alpha^1 \int_{\hat{\theta}^{I}_2}^{\hat{\theta}^U_2} f^1(x)dx -L^h_2 \alpha^0 \int_{\hat{\theta}^{I}_2}^{\hat{\theta}^U_2} f^0(x)dx \Big\} \\
        &\qquad \geq N_2\Big \{ L^h_1 \alpha^1 \int_{LB_1}^{\hat{\theta}_1} f^1(x)dx -L^h_2 \alpha^0 \int_{LB_1}^{\hat{\theta}_1} f^0(x)dx \Big\} 
    \end{align*}
Where the inequality is based on Theorem~\ref{thm:theta} such that $\hat{\theta}^b_2 \geq \hat{\theta}^{int}_2 \geq \hat{\theta}_1 \geq LB_1$, and our assumption that the differences in mode between $f^1$ and $f^0$. Therefore, by combining all the pieces together, we could deduce the following conclusion when the condition is satisfied:
    $\mathbb{E}[\mathcal{L}(I, \hat{f}_1^0, \hat{f}_1^1)] \leq \mathbb{E}[\mathcal{L}(U, \hat{f}_1^0, \hat{f}_1^1)]$
\section{Details on Numerical Experiment Setups}\label{sec:more-experiments}

\textbf{Parameter descriptions on real-world data experiments}: For the \emph{Adult} dataset the true underlying distributions were estimated to be Beta distributions with parameters Beta(1.94, 3.32) and Beta(1.13, 4.99) for group $a$ (White) label 1 and 0, respectively, and Beta(1.97, 3.53) and Beta(1.19, 6.10) for group $b$ (non-White) label 1 and 0, respectively. We used 2.5\% of the data to fit initial assumed distributions Beta(1.83, 3.32) and Beta(1.22, 4.99) for group $a$ label 1 and 0, respectively, and Beta(1.74, 3.53) and Beta(1.28, 6.10) for group $b$ label 1 and 0, respectively. The equality of opportunity fairness constraint is imposed throughout. The exploration frequency $\{\epsilon_t\}$ is reduced with the fixed schedule of being subtracted by 0.1 after observing every 10000 samples. 

For the \emph{FICO} dataset, the true underlying distributions were estimated to be Beta distributions with parameters Beta(2.16, 1.27) and Beta(1.06, 3.98) for group $a$ (White) label 1 and 0, respectively, and Beta(1.71, 1.62) and Beta(1.16, 5.51) for group $b$ (non-White) label 1 and 0, respectively. We used 0.3\% of the data to fit initial assumed distributions Beta(2.34, 1.27) and Beta(1.01, 3.98) for group $a$ label 1 and 0, respectively, and Beta(1.98, 1.62) and Beta(1.42, 5.51) for group $b$ label 1 and 0, respectively. The equality of opportunity fairness constraint is imposed. The exploration frequency $\{\epsilon_t\}$ is reduced with the fixed schedule of being subtracted by 0.1 after observing every 17000 samples. 

For the \emph{Retiring Adult} dataset, the true underlying distributions were estimated to be Beta distributions with parameters Beta(2.83, 2.16) and Beta(1.22, 2.57) for group $a$ (White) label 1 and 0, respectively, and Beta(2.30, 2.53) and Beta(1.03, 3.30) for group $b$ (non-White) label 1 and 0, respectively. We used 3\% of the data to fit initial assumed distributions Beta(2.70, 2.16) and Beta(1.28, 2.57) for group $a$ label 1 and 0, respectively, and Beta(2.22, 2.53) and Beta(1.16, 3.30) for group $b$ label 1 and 0, respectively. The equality of opportunity fairness constraint is imposed. The exploration frequency $\{\epsilon_t\}$ is reduced with the fixed schedule of being subtracted by 0.1 after observing every 100k samples.

\section{Debiasing with two unknown parameters: a Gaussian distribution with two unknown parameters mean $\mu$ and variance $\sigma^2$}\label{app:two_parameters}

In this section, we extend our algorithm to debias the estimates of distributions with two unknown parameters. Specifically, we consider a single group, and assume that the underlying feature-label distributions are Gaussian distributions for which both the mean and variance are potentially incorrectly estimated. 

For this experiment only, we follow our \texttt{active debiasing} algorithm, with a choice of medians as reference points (i.e., $\tau^i=50, \forall i$), and setting the thresholds UB (See Definition~\ref{def:UB}) and LB so that the reference points are the medians of the truncated distribution between the bounds and the classifier $\theta$. We then follow Algorithm~\ref{def:db-alg}'s procedure with the same type of exploitation and exploration decisions, and with the additional step that now we update both parameters when updating the underlying estimates. 

In order to update the mean and  variance estimates for obtaining $\hat{f}^i_{t}$, we find the sample mean and sample variance of the collected data, incrementally. However, we note that the obtained sample mean and sample variances are \emph{for truncated distributions}; the truncations are due to the presence of a classifier which limits the admission of a samples, as well as due to our proposed bounds UB and LB in the data collection procedure. We therefore need to convert between the estimated statistics for the truncated distribution and those of the full distribution accordingly. 

Specifically, we obtain the sample mean of the truncated distribution as follows: 
\begin{align*}
\hat{\mu}^i_{t+1}=\frac{x_1+x_2+...+x_{n_i}+x^\dagger}{N^i_t+1} =  \frac{N^i_t}{N^i_t+1}\hat{\mu}^i_{t}+\frac{x^\dagger}{N^i_t+1}, \quad i \in \{0, 1\}~.
\end{align*}
where $N^i_t$ is the existing number of agents in the pool, and $\mu^i_t$ is the current (truncated) mean value estimate for label $i=\{0, 1\}$.

For the sample (truncated) variance for group $i$, $(\hat{s}^i_t)^2$, the updating procedure is:
\begin{align*}
(\hat{s}^i_{t+1})^2&=\frac{\sum_{j=1}^{N^i_t}(\hat{\mu}^i_{t}-x_j)^2+(\hat{\mu}^i_{t}-x^\dagger)^2}{N^i_t+1-1} \\ &=\frac{\sum_{j=1}^{N^i_t} x^2_j+(x^\dagger)^2-(N^i_t+1)(\hat{\mu}^i_{t})^2}{N^i_t+1-1} = \frac{N^i_t-1}{N^i_t} (\hat{s}^i_{t})^2+\frac{(x^\dagger)^2-(\hat{\mu}^i_{t})^2}{N^i_t}, \quad i \in \{0,1\}~.
\end{align*}

After finding the above estimates of the mean and variance of the truncated distribution, we need to estimate the mean and variance of the \emph{full} underlying distribution. We first note that given our choice of bounds UB and LB, the mean of the underlying distribution is (assumed to be) the same as that of the truncated distribution. To find the untruncated variance for the full distribution, we use the following relation between the variances of truncated and untruncated Gaussian distributions:
\begin{align*}
    Var(x|a\leq x\leq b) =s^2 = \sigma^2 \Bigg[1+\frac{\alpha\phi(\alpha)-\beta\phi(\beta)}{\Phi(\beta)-\Phi(\alpha)}-(\frac{\phi(\alpha)-\phi(\beta)}{\Phi(\beta)-\Phi(\alpha)})^2\Bigg]
\end{align*}
where $\alpha =\frac{a-\mu}{\sigma}$, $\beta = \frac{b-\mu}{\sigma}$, $\phi(x)=\frac{1}{\sqrt{2\pi}} e^{-\frac{1}{2}x^2}$ and $\Phi(x)=\frac{1}{2}(1+erf(\frac{x}{\sqrt{2}}))$. In our algorithm, $a=UB$ and $b=\theta$ for $i=1$, and $a=\theta$ and $b=LB$ for $i=0$. We note that in both cases, we can drop the third term in the above formula since based on our algorithm, $a,b$ are symmetric around the mean value, so that $\phi(\alpha)=\phi(\beta)$. We solve the above equations to find $\hat{\sigma}^i_t$ from the truncated estimates $\hat{s}^i_t$. 

\begin{figure}[t]
	\centering
	\subfigure[Debiasing the means.]
	{
		\includegraphics[width=0.4\textwidth]{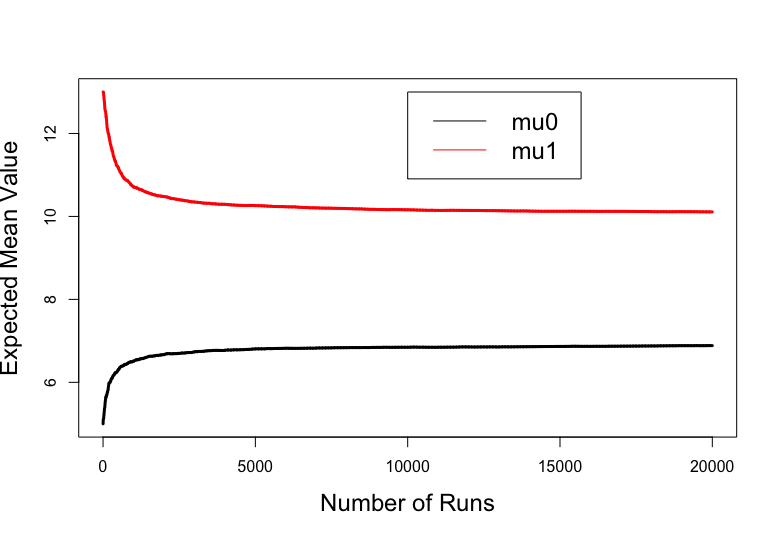}
		\label{fig:fix-mu}
	}%
	\subfigure[Debiasing the variances.]
	{
		\includegraphics[width=0.4\textwidth]{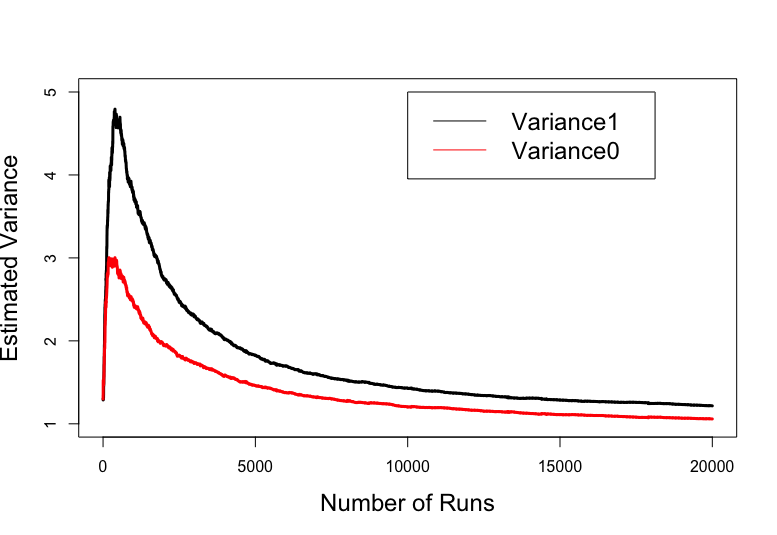}
		\label{fig:fix-sigma}
	}%
	\caption{Debiasing algorithm when both mean and variance of a Gaussian distribution are incorrectly estimated. The true underlying distributions are $f^1\sim N(10,1)$ and $f^0\sim N(7,1)$, and the initial estimates are $\hat{f}^1_0\sim N(13,1.3)$ and $\hat{f}^0_0\sim N(5,1.3)$. The algorithm corrects both biases in the long run.}%
	\label{fig:two-parameter-debiasing}
\end{figure}

Figure~\ref{fig:two-parameter-debiasing} shows that the debiasing algorithm with the update procedures described above can debias both parameters in the long run. We do observe that the debiasing of the variance initially increases its error. This is because, initially, when observing samples outside of its believed range (due to a combination of incorrectly estimated means and variances), the algorithm increases its estimates of the variance to explain such samples. However, as the estimate of the mean is corrected, the variance can be reduced as well and become consistent with the collected observations. Ultimately, both parameters will be correctly estimated. 

\clearpage
\renewcommand{\thesection}{\Alph{section}}
\setcounter{section}{0}

\end{document}